\documentclass{article}

\usepackage{times}
\usepackage{graphicx} 
\usepackage{subfigure}
\usepackage{amsmath}
\usepackage{amssymb}

\usepackage{natbib}

\usepackage{algorithm}
\usepackage{algorithmic}

\usepackage{hyperref}

\renewcommand{\v}[1]{\mathbf{#1}}

\newcommand{\betavae}{\ensuremath{\beta}-VAE\space}
\newcommand{\betavaedae}{\ensuremath{\beta\text{-VAE}_{DAE}}\space}

\newcommand\cut[1]{}

\usepackage[accepted]{icml2017}

\icmltitlerunning{DARLA: Improving Zero-Shot Transfer in Reinforcement Learning}

\begin{document}

\twocolumn[
\icmltitle{DARLA: Improving Zero-Shot Transfer in Reinforcement Learning}



\icmlsetsymbol{equal}{*}

\begin{icmlauthorlist}
\icmlauthor{Irina~Higgins}{equal,dm}
\icmlauthor{Arka~Pal}{equal,dm}
\icmlauthor{Andrei~Rusu}{dm}
\icmlauthor{Loic~Matthey}{dm}
\icmlauthor{Christopher~Burgess}{dm}
\icmlauthor{Alexander~Pritzel}{dm}
\icmlauthor{Matthew~Botvinick}{dm}
\icmlauthor{Charles~Blundell}{dm}
\icmlauthor{Alexander~Lerchner}{dm}
\end{icmlauthorlist}

\icmlaffiliation{dm}{DeepMind, 6 Pancras Square, Kings Cross, London, N1C 4AG, UK}

\icmlcorrespondingauthor{Irina Higgins}{irinah@google.com}
\icmlcorrespondingauthor{Arka Pal}{arkap@google.com}

\icmlkeywords{transfer learning, domain adaptation, reinforcement learning, disentangled representations, beta-VAE}

\vskip 0.3in
]



\printAffiliationsAndNotice{\icmlEqualContribution} 

\begin{abstract}
Domain adaptation is an important open problem in deep reinforcement learning (RL). In many scenarios of interest data is hard to obtain, so agents may learn a \emph{source} policy in a setting where data is readily available, with the hope that it generalises well to the \emph{target} domain. We propose a new multi-stage RL agent, DARLA (DisentAngled Representation Learning Agent), which learns to see before learning to act. DARLA's vision is based on learning a disentangled representation of the observed environment. Once DARLA can see, it is able to acquire source policies that are robust to many domain shifts - even with no access to the target domain. DARLA significantly outperforms conventional baselines in zero-shot domain adaptation scenarios, an effect that holds across a variety of RL environments (Jaco arm, DeepMind Lab) and base RL algorithms (DQN, A3C and EC).
\end{abstract}

\section{Introduction}
\label{sec_introduction}
Autonomous agents can learn how to maximise future expected rewards by choosing how to act based on incoming sensory observations via reinforcement learning (RL). Early RL approaches did not scale well to environments with large state spaces and high-dimensional raw observations \citep{Sutton_Barto_1998}. A commonly used workaround was to embed the observations in a lower-dimensional space, typically via hand-crafted and/or privileged-information features. Recently, the advent of deep learning and its successful combination with RL has enabled end-to-end learning of such embeddings directly from raw inputs, sparking success in a wide variety of previously challenging RL domains \citep{Mnih_etal_2015, Mnih_etal_2016, Jaderberg_etal_2017}. Despite the seemingly universal efficacy of deep RL, however, fundamental issues remain. These include data inefficiency, the reactive nature and general brittleness of learnt policies to changes in input data distribution, and lack of model interpretability \citep{Garnelo_etal_2016, Lake_etal_2016}. This paper focuses on one of these outstanding issues: the ability of RL agents to deal with changes to the input distribution, a form of transfer learning known as \textit{domain adaptation} \citep{Bengio_etal_2013}. In domain adaptation scenarios, an agent trained on a particular input distribution with a specified reward structure (termed the \textit{source domain}) is placed in a setting where the input distribution is modified but the reward structure remains largely intact (the \textit{target domain}). We aim to develop an agent that can learn a robust policy using observations and rewards obtained exclusively within the source domain. Here, a policy is considered as robust if it generalises with minimal drop in performance to the target domain without extra fine-tuning.

Past attempts to build RL agents with strong domain adaptation performance highlighted the importance of learning good internal representations of raw observations \citep{Finn_etal_2015, Raffin_etal_2017, Pan_Yang_2009, Barretto_et_al_2016, Littman_etal_2001}. Typically, these approaches tried to align the source and target domain representations by utilising observation and reward signals from both domains \citep{Tzeng_etal_2016, Daftry_etal_2016, Parisotto_etal_2015, Guez_etal_2012, Talvitie_Singh_2007, Niekum_etal_2013, Gupta_etal_2017, Finn_etal_2017, Rajendran_etal_2017}. In many scenarios, such as robotics, this reliance on target domain information can be problematic, as the data may be expensive or difficult to obtain \citep{Finn_etal_2017, Rusu_etal_2016b}. Furthermore, the target domain may simply not be known in advance. On the other hand, policies learnt exclusively on the source domain using existing deep RL approaches that have few constraints on the nature of the learnt representations often overfit to the source input distribution, resulting in poor domain adaptation performance \citep{Lake_etal_2016, Rusu_etal_2016b}.

\begin{figure}[t]
\begin{center}
\centerline{\includegraphics[width=0.7\columnwidth]{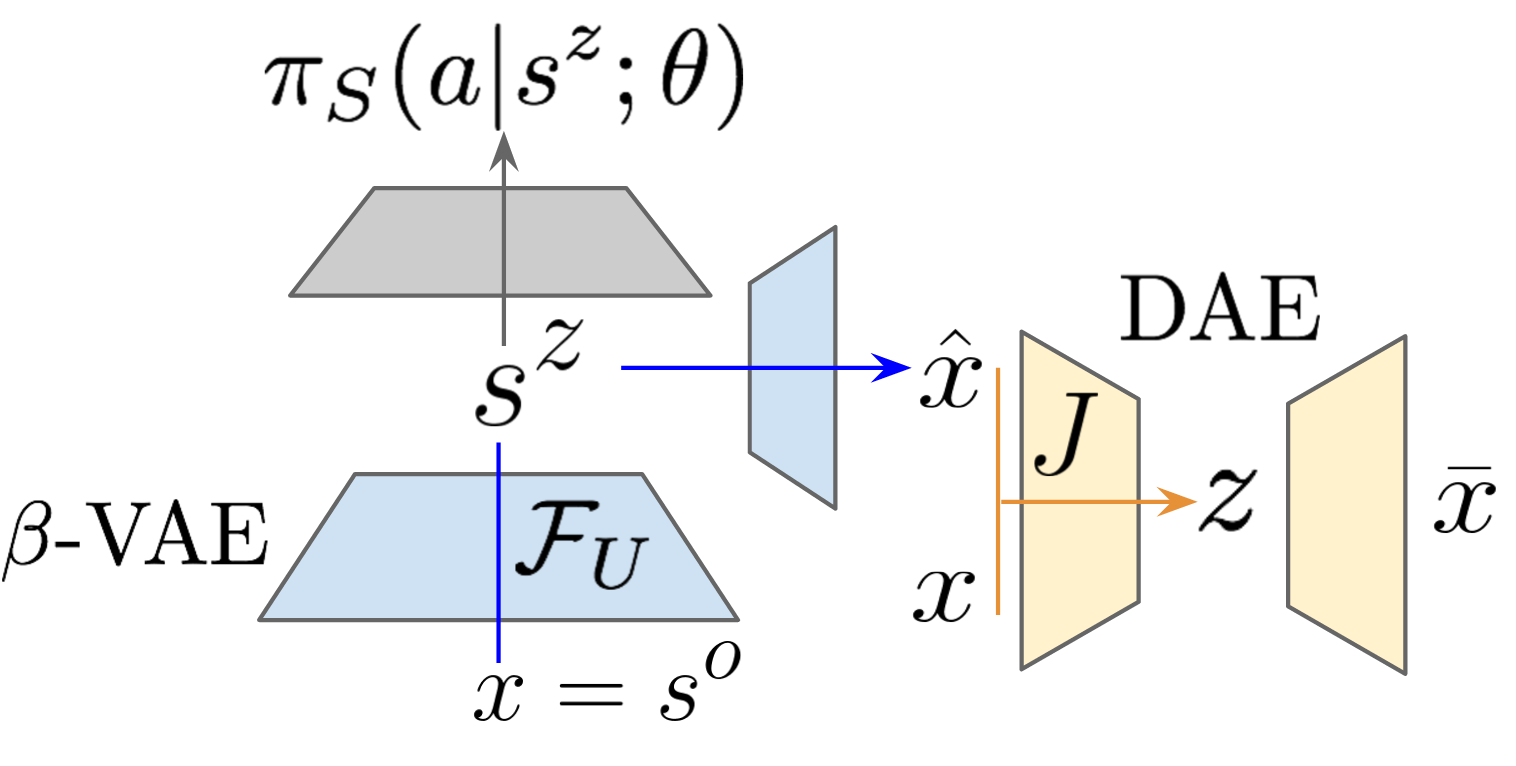}}
\caption{Schematic representation of DARLA. Yellow represents the denoising autoencoder part of the model, blue represents the \betavae part of the model, and grey represents the policy learning part of the model.}
\label{fig_vae_ae_agent}
\end{center}
\vspace{-30pt}
\end{figure}

We propose tackling both of these issues by focusing instead on learning representations which capture an underlying low-dimensional factorised representation of the world and are therefore not task or domain specific. Many naturalistic domains such as video game environments, simulations and our own world are well described in terms of such a structure. Examples of such factors of variation are object properties like colour, scale, or position; other examples correspond to general environmental factors, such as geometry and lighting. We think of these factors as a set of high-level parameters that can be used by a world graphics engine to generate a particular natural visual scene \citep{Kulkarni_etal_2015}. Learning how to project raw observations into such a factorised description of the world is addressed by the large body of literature on disentangled representation learning \citep{Schmidhuber_1992, Desjardins_etal_2012, Cohen_Welling_2014, Cohen_Welling_2015, Kulkarni_etal_2015, Hinton_etal_2011, Rippel_Adams_2013, Reed_etal_2014, Yang_etal_2015, Goroshin_etal_2015, Kulkarni_etal_2015, Cheung15, Whitney_etal_2016, Karaletsos_etal_2016, Chen_etal_2016, Higgins_et_al_2017}. Disentangled representations are defined as interpretable, factorised latent representations where either a single latent or a group of latent units are sensitive to changes in single ground truth factors of variation used to generate the visual world, while being invariant to changes in other factors \citep{Bengio_etal_2013}. The theoretical utility of disentangled representations for supervised and reinforcement learning has been described before \citep{Bengio_etal_2013, Higgins_et_al_2017, Ridgeway2016-nj}; however, to our knowledge, it has not been empirically validated to date.

We demonstrate how disentangled representations can improve the robustness of RL algorithms in domain adaptation scenarios by introducing \emph{DARLA} (DisentAngled Representation Learning Agent), a new RL agent capable of learning a robust policy on the source domain that achieves significantly better out-of-the-box performance in domain adaptation scenarios compared to various baselines. DARLA relies on learning a latent state representation that is shared between the source and target domains, by learning a disentangled representation of the environment's generative factors. Crucially, DARLA does not require target domain data to form its representations. Our approach utilises a three stage pipeline: 1) learning to see, 2) learning to act, 3) transfer. During the first stage, DARLA develops its vision, learning to parse the world in terms of basic visual concepts, such as objects, positions, colours, etc. by utilising a stream of raw unlabelled observations -- not unlike human babies in their first few months of life \citep{Leat_etal_2009, Candy_etal_2009}. In the second stage, the agent utilises this disentangled visual representation to learn a robust source policy. In stage three, we demonstrate that the DARLA source policy is more robust to domain shifts, leading to a significantly smaller drop in performance in the target domain even when no further policy finetuning is allowed (median 270.3\% improvement). These effects hold consistently across a number of different RL environments \citep[DeepMind Lab and Jaco/MuJoCo:][]{deepmind_lab, Todorov_etal_2012} and algorithms \citep[DQN, A3C and Episodic Control:][]{Mnih_etal_2015, Mnih_etal_2016, Blundell_etal_2016}.

\section{Framework}
\subsection{Domain adaptation in Reinforcement Learning}
We now formalise domain adaptation scenarios in a reinforcement learning (RL) setting. We denote the source and target domains as $D_S$ and $D_T$, respectively. Each domain corresponds to an MDP defined as a tuple $D_S \equiv (\mathcal{S}_S, \mathcal{A}_S, \mathcal{T}_S, R_S)$ or $D_T \equiv (\mathcal{S}_T, \mathcal{A}_T, \mathcal{T}_T, R_T)$ (we assume a shared fixed discount factor $\gamma$), each with its own state space $\mathcal{S}$, action space $\mathcal{A}$, transition function $\mathcal{T}$ and reward function $R$.\footnote{For further background on the notation relating to the RL paradigm, see Section~\ref{sec_reinforcement_learning_appendix} in the Supplementary Materials.} In domain adaptation scenarios the states $\mathcal{S}$ of the source and the target domains can be quite different, while the action spaces $\mathcal{A}$ are shared and the transitions $\mathcal{T}$ and reward functions $R$ have structural similarity. For example, consider a domain adaptation scenario for the Jaco robotic arm, where the MuJoCo \citep{Todorov_etal_2012} simulation of the arm is the source domain, and the real world setting is the target domain. The state spaces (raw pixels) of the source and the target domains differ significantly due to the perceptual-reality gap \citep{Rusu_etal_2016b}; that is to say, $\mathcal{S}_S \neq \mathcal{S}_T$. Both domains, however, share action spaces ($\mathcal{A}_S = \mathcal{A}_T$), since the policy learns to control the same set of actuators within the arm. Finally, the source and target domain transition and reward functions share structural similarity ($\mathcal{T}_S \approx \mathcal{T}_T$ and $R_S \approx R_T$), since in both domains transitions between states are governed by the physics of the world and the performance on the task depends on the relative position of the arm's end effectors (i.e. fingertips) with respect to an object of interest.

\subsection{DARLA}
\label{sec_darla}
In order to describe our proposed DARLA framework, we assume that there exists a set $\mathcal{M}$ of MDPs that is the set of all \emph{natural world} MDPs, and each MDP $D_i$ is sampled from $\mathcal{M}$. We define $\mathcal{M}$ in terms of the state space $\hat{\mathcal{S}}$ that contains all possible conjunctions of high-level factors of variation necessary to generate any naturalistic observation in any $D_i \in \mathcal{M}$. A \emph{natural world} MDP $D_i$ is then one whose state space $\mathcal{S}$ corresponds to some subset of $\hat{\mathcal{S}}$. In simple terms, we assume that there exists \emph{some} shared underlying structure between the MDPs $D_i$ sampled from $\mathcal{M}$. We contend that this is a reasonable assumption that permits inclusion of many interesting problems, including being able to characterise our own reality \citep{Lake_etal_2016}.

We now introduce notation for two state space variables that may in principle be used interchangeably within the source and target domain MDPs $D_S$ and $D_T$ -- the agent observation state space $\mathcal{S}^o$, and the agent's internal latent state space $\mathcal{S}^z$.\footnote{Note that we do not assume these to be Markovian i.e. it is not necessarily the case that $p(s^o_{t+1}|s^o_t) = p(s^o_{t+1}|s^o_t, s^o_{t-1}, \ldots, s^o_1)$, and similarly for $s^z$. Note the index $t$ here corresponds to time.} $\mathcal{S}^o_i$ in $D_i$ consists of raw (pixel) observations $s^o_i$ generated by the true world simulator from a sampled set of data generative factors $\hat{s}_i$, i.e. $s^o_i \sim \bf{Sim}(\hat{s}_i)$. $\hat{s}_i$ is sampled by some distribution or process $\mathcal{G}_i$ on $\hat{\mathcal{S}}$, $\hat{s}_i \sim \mathcal{G}_i(\hat{\mathcal{S}})$. 

Using the newly introduced notation, domain adaptation scenarios can be described as having different sampling processes $\mathcal{G}_S$ and $\mathcal{G}_T$ such that $\hat{s}_S \sim \mathcal{G}_S(\hat{\mathcal{S}})$ and $\hat{s}_T \sim \mathcal{G}_T(\hat{\mathcal{S}})$ for the source and target domains respectively, and then using these to generate different agent observation states $s^o_S \sim \bf{Sim}(\hat{s}_S)$ and $s^o_T \sim \bf{Sim}(\hat{s}_T)$. Intuitively, consider a source domain where oranges appear in blue rooms and apples appear in red rooms, and a target domain where the object/room conjunctions are reversed and oranges appear in red rooms and apples appear in blue rooms. While the true data generative factors of variation $\hat{\mathcal{S}}$ remain the same - room colour (blue or red) and object type (apples and oranges) - the particular source and target distributions $\mathcal{G}_S$ and $\mathcal{G}_T$ differ.

Typically deep RL agents \citep[e.g. ][]{Mnih_etal_2015, Mnih_etal_2016} operating in an MDP $D_i \in \mathcal{M}$ learn an end-to-end mapping from raw (pixel) observations $s^o_i \in \mathcal{S}^o_i$ to actions $a_i \in \mathcal{A}_i$ (either directly or via a value function $Q_i(s^o_i, a_i)$ from which actions can be derived). In the process of doing so, the agent implicitly learns a function $\mathcal{F}: \mathcal{S}^o_i \rightarrow \mathcal{S}^z_i$ that maps the typically high-dimensional raw observations $s^o_i$ to typically low-dimensional latent states $s^z_i$; followed by a policy function $\pi_i: \mathcal{S}^z_i \rightarrow \mathcal{A}_i$ that maps the latent states $s^z_i$ to actions $a_i \in \mathcal{A}_i$. In the context of domain adaptation, if the agent learns a naive latent state mapping function $\mathcal{F}_S:  \mathcal{S}^o_S \rightarrow  \mathcal{S}^z_S$ on the source domain using reward signals to shape the representation learning, it is likely that $\mathcal{F}_S$ will overfit to the source domain and will not generalise well to the target domain. Returning to our intuitive example, imagine an agent that has learnt a policy to pick up oranges and avoid apples on the source domain. Such a source policy $\pi_S$ is likely to be based on an \emph{entangled} latent state space $\mathcal{S}^z_S$ of object/room conjunctions: {oranges/blue} $\rightarrow$ {good}, {apples/red} $\rightarrow$ {bad}, since this is arguably the most efficient representation for maximising expected rewards on the source task in the absence of extra supervision signals suggesting otherwise. A source policy $\pi_S(a|s^z_S; \theta)$ based on such an entangled latent representation $s^z_S$ will not generalise well to the target domain without further fine-tuning, since $\mathcal{F}_S(s^o_S) \neq \mathcal{F}_S(s^o_T)$ and therefore crucially $S^z_S \neq S^z_T$.

On the other hand, since both $\hat{s}_S \sim \mathcal{G}_S(\hat{\mathcal{S}})$ and $\hat{s}_T \sim \mathcal{G}_T(\hat{\mathcal{S}})$ are sampled from the same \emph{natural world} state space $\hat{\mathcal{S}}$ for the source and target domains respectively, it should be possible to learn a latent state mapping function 
$\hat{\mathcal{F}}: \mathcal{S}^o \rightarrow \mathcal{S}^z_{\hat{\mathcal{S}}}$, which projects the agent observation state space $\mathcal{S}^o$ to a latent state space $\mathcal{S}^z_{\hat{\mathcal{S}}}$ expressed in terms of factorised data generative factors that are representative of the natural world i.e. $S^z_{\hat{S}} \approx \hat{S}$. Consider again our intuitive example, where $\hat{\mathcal{F}}$ maps agent observations ($s^o_S$: orange in a blue room) to a factorised or \emph{disentangled} representation expressed in terms of the data generative factors ($s^z_{\hat{\mathcal{S}}}$: \emph{room type} = blue; \emph{object type} = orange). Such a \emph{disentangled} latent state mapping function should then directly generalise to both the source and the target domains, so that $\hat{\mathcal{F}}(s^o_S) = \hat{\mathcal{F}}(s^o_T) = s^z_{\hat{\mathcal{S}}}$. Since $\mathcal{S}^z_{\hat{\mathcal{S}}}$ is a \emph{disentangled} representation of object and room attributes, the source policy $\pi_S$ can learn a decision boundary that ignores the irrelevant room attributes: {oranges} $\rightarrow$ {good}, {apples} $\rightarrow$ {bad}. Such a policy would then generalise well to the target domain out of the box, since  $\pi_S(a|\hat{\mathcal{F}}(s^o_S); \theta) = \pi_T(a|\hat{\mathcal{F}}(s^o_T); \theta) = \pi_T(a|s^z_{\hat{\mathcal{S}}}; \theta)$. Hence, DARLA is based on the idea that a good quality $\hat{\mathcal{F}}$ learnt exclusively on the source domain $D_S \in \mathcal{M}$ will zero-shot-generalise to all target domains $D_i \in \mathcal{M}$, and therefore the source policy $\pi(a|\mathcal{S}^z_{\hat{\mathcal{S}}}; \theta)$ will also generalise to all target domains $D_i \in \mathcal{M}$ out of the box.

Next we describe each of the stages of the DARLA pipeline that allow it to learn source policies $\pi_S$ that are robust to domain adaptation scenarios, despite being trained with no knowledge of the target domains (see Fig.~\ref{fig_vae_ae_agent} for a graphical representation of these steps):

\emph{1) Learn to see} (unsupervised learning of $\mathcal{F}_U$) -- the task of inferring a factorised set of generative factors $\mathcal{S}^z_{\hat{\mathcal{S}}} = \hat{S}$ from observations $\mathcal{S}^o$ is the goal of the extensive disentangled factor learning literature \citep[e.g. ][]{Chen_etal_2016, Higgins_et_al_2017}. Hence, in stage one we learn a mapping $\mathcal{F}_U: \mathcal{S}^o_U \rightarrow \mathcal{S}^z_U$, where $\mathcal{S}^z_U \approx \mathcal{S}^z_{\hat{\mathcal{S}}}$ ($U$ stands for `unsupervised') using an unsupervised model for learning disentangled factors that utilises observations collected by an agent with a random policy $\pi_U$ from a visual pre-training MDP $D_U \in \mathcal{M}$. Note that we require sufficient variability of factors and their conjunctions in $D_U$ in order to have $S^z_U \approx \mathcal{S}^z_{\hat{\mathcal{S}}}$;

\emph{2) Learn to act} (reinforcement learning of $\pi_S$ in the source domain $D_S$ utilising previously learned $\mathcal{F}_U$) -- an agent that has learnt to see the world in stage one in terms of the natural data generative factors is now exposed to a source domain $D_S \in \mathcal{M}$. The agent is tasked with learning the source policy $\pi_S(a|s^z_S; \theta)$, where $s^z_S = \mathcal{F}_U(s^o_S) \approx s^z_{\hat{\mathcal{S}}}$, via a standard reinforcement learning algorithm. Crucially, we do not allow $\mathcal{F}_U$ to be modified (e.g. by gradient updates) during this phase;

\emph{3) Transfer} (to a target domain $D_T$) -- in the final step, we test how well the policy $\pi_S$ learnt on the source domain generalises to the target domain $D_T \in \mathcal{M}$ in a zero-shot domain adaptation setting, i.e. the agent is evaluated on the target domain without retraining. We compare the performance of policies learnt with a disentangled latent state $\mathcal{S}^z_{\hat{\mathcal{S}}}$ to various baselines where the latent state mapping function $\mathcal{F}_U$ projects agent observations $s^o$ to entangled latent state representations $s^z$.

\subsection{Learning disentangled representations}
\label{sec_disent_rep}
In order to learn $\mathcal{F}_U$, DARLA utilises \betavae \ \citep{Higgins_et_al_2017}, a state-of-the-art unsupervised model for automated discovery of factorised latent representations from raw image data. \betavae \ is a modification of the variational autoencoder framework \citep{Kingma_Welling_2014, Rezende_etal_2014} that controls the nature of the learnt latent representations by introducing an adjustable hyperparameter $\beta$ to balance reconstruction accuracy with latent channel capacity and independence constraints. It maximises the objective:

\setlength{\belowdisplayskip}{3pt} \setlength{\belowdisplayshortskip}{3pt}
\setlength{\abovedisplayskip}{-12pt} \setlength{\abovedisplayshortskip}{-12pt}

\begin{align}
\mathcal{L}(\theta, \phi; \v{x}, \v{z}, \beta)  =&  \mathbb{E}_{q_\phi(\v{z}|\v{x})} [\log p_\theta(\v{x} | \v{z})] \nonumber \\
    & - \beta \ D_{KL}(q_\phi(\v{z}|\v{x}) || p(\v{z})) 
 \label{eq_beta_vae}
\end{align}
where $\phi$, $\theta$ parametrise the distributions of the encoder and the decoder respectively. Well-chosen values of $\beta$ - usually larger than one ($\beta>1$) - typically result in more disentangled latent representations $\v{z}$ by limiting the capacity of the latent information channel, and hence encouraging a more efficient factorised encoding through the increased pressure to match the isotropic unit Gaussian prior $p(\v{z})$ \citep{Higgins_et_al_2017}.

\subsubsection{Perceptual Similarity Loss}
\label{sec_perceptual_similarity_loss}
The cost of increasing $\beta$ is that crucial information about the scene may be discarded in the latent representation $\v{z}$, particularly if that information takes up a small proportion of the observations $\v{x}$ in pixel space. We encountered this issue in some of our tasks, as discussed in Section~\ref{sec_tasks_lab}. The shortcomings of calculating the log-likelihood term $\mathbb{E}_{q_\phi(\v{z}|\v{x})}[\log p_\theta(\v{x}|\v{z})]$ on a per-pixel basis are known and have been addressed in the past by calculating the reconstruction cost in an abstract, high-level feature space given by \textit{another} neural network model, such as a GAN \citep{Goodfellow_etal_2014} or a pre-trained AlexNet \citep{Krizhevsky_etal_2012, Larsen_etal_2016, Dosovitskiy_Brox_2016, Warde-Farley_Bengio_2017}. In practice we found that pre-training a denoising autoencoder \citep{Vincent_etal_2015} on data from the visual pre-training MDP $D_U \in \mathcal{M}$ worked best as the reconstruction targets for \betavae \ to match (see Fig.~\ref{fig_vae_ae_agent} for model architecture and Sec.~\ref{sec_sup_ae} in Supplementary Materials for implementation details). The new \betavaedae model was trained according to Eq.~\ref{eq_beta_vae_ae}:

\begin{align}
\mathcal{L}(\theta, \phi; \v{x}, \v{z}, \beta)  = & - \mathbb{E}_{q_\phi(\v{z}|\v{x})}\left \|J(\hat{\v{x}}) - J(\v{x})  \right \|_2^2   \nonumber \\
 &- \beta \ D_{KL}(q_\phi(\v{z}|\v{x}) || p(\v{z}))
\label{eq_beta_vae_ae}   
\end{align}

\setlength{\belowdisplayskip}{10pt} \setlength{\belowdisplayshortskip}{10pt}
\setlength{\abovedisplayskip}{10pt} \setlength{\abovedisplayshortskip}{10pt}

where $\hat{\v{x}} \sim  p_\theta(\v{x} | \v{z})$ and $J : \mathbb{R}^{W \times H \times C} \rightarrow  \mathbb{R}^N$ is the function that maps images from pixel space with dimensionality $W \times H \times C$ to a high-level feature space with dimensionality $N$ given by a stack of pre-trained DAE layers up to a certain layer depth. Note that by replacing the pixel based reconstruction loss in Eq.~\ref{eq_beta_vae} with high-level feature reconstruction loss in Eq.~\ref{eq_beta_vae_ae} we are no longer optimising the variational lower bound, and \betavaedae with $\beta=1$ loses its equivalence to the Variational Autoencoder (VAE) framework as proposed by \cite{Kingma_Welling_2014, Rezende_etal_2014}. In this setting, the only way to interpret $\beta$ is as a mixing coefficient that balances the capacity of the latent channel $\v{z}$ of \betavaedae against the pressure to match the high-level features within the pre-trained DAE.

\subsection{Reinforcement Learning Algorithms}
\label{sec_reinforcement_learning_algorithms}
We used various RL algorithms \citep[DQN, A3C and Episodic Control:][]{Mnih_etal_2015, Mnih_etal_2016, Blundell_etal_2016} to learn the source policy $\pi^S$ during stage two of the pipeline using the latent states $s^z$ acquired by \betavae \ based models during stage one of the DARLA pipeline. 

\emph{Deep Q Network}
(DQN) \citep{Mnih_etal_2015} is a variant of the Q-learning algorithm \citep{Watkins_1989} that utilises deep learning. It uses a neural network to parametrise an approximation for the action-value function $Q(s, a; \theta)$ using parameters $\theta$. 
 
\emph{Asynchronous Advantage Actor-Critic}
(A3C) \citep{Mnih_etal_2016} is an asynchronous implementation of the advantage actor-critic paradigm \citep{Sutton_Barto_1998, Degris_etal_2012}, where separate threads run in parallel and perform updates to shared parameters. The different threads each hold their own instance of the environment and have different exploration policies, thereby decorrelating parameter updates without the need for experience replay. Therefore, A3C is an online algorithm, whereas DQN learns its policy offline, resulting in different learning dynamics between the two algorithms.

\emph{Model-Free Episodic Control}
(EC) \citep{Blundell_etal_2016} was proposed as a complementary learning system to the other RL algorithms described above. The EC algorithm relies on near-determinism of state transitions and rewards in RL environments; in settings where this holds, it can exploit these properties to memorise which action led to high returns in similar situations in the past. Since in its simplest form EC relies on a lookup table, it learns good policies much faster than value-function-approximation based deep RL algorithms like DQN trained via gradient descent - at the cost of generality (i.e. potentially poor performance in non-deterministic environments).

We also compared our approach to that of \emph{UNREAL} \citep{Jaderberg_etal_2017}, a recently proposed RL algorithm which also attempts to utilise unsupervised data in the environment. The UNREAL agent takes as a base an LSTM A3C agent \citep{Mnih_etal_2016} and augments it with a number of unsupervised auxiliary tasks that make use of the rich perceptual data available to the agent besides the (sometimes very sparse) extrinsic reward signals. This auxiliary learning tends to improve the representation learnt by the agent. See Sec.~\ref{sec_background_on_rl_algorithms} in Supplementary Materials for further details of the algorithms above.

\section{Tasks}
\label{sec_tasks}
We evaluate the performance of DARLA on different task and environment setups that probe subtly different aspects of domain adaptation. As a reminder, in Sec.~\ref{sec_darla} we defined $\hat{\mathcal{S}}$ as a state space that contains all possible conjunctions of high-level factors of variation necessary to generate any naturalistic observation in any $D_i \in \mathcal{M}$. During domain adaptation scenarios agent observation states are generated according to $s^o_S \sim \bf{Sim}_S(\hat{s}_S)$ and $s^o_T \sim \bf{Sim}_T(\hat{s}_T)$ for the source and target domains respectively, where $\hat{s}_S$ and $\hat{s}_T$ are sampled by some distributions or processes $\mathcal{G}_S$ and $\mathcal{G}_T$ according to  $\hat{s}_S \sim \mathcal{G}_S(\hat{\mathcal{S}})$ and $\hat{s}_T \sim \mathcal{G}_T(\hat{\mathcal{S}})$.

We use DeepMind Lab \citep{deepmind_lab} to test a version of domain adaptation setup where the source and target domain observation simulators are equal ($\bf{Sim}_S = \bf{Sim}_T$), but the processes used to sample  $\hat{s}_S$ and  $\hat{s}_T$ are different ($\mathcal{G}_S \neq \mathcal{G}_T$). We use the Jaco arm with a matching MuJoCo simulation environment \citep{Todorov_etal_2012} in two domain adaptation scenarios: simulation to simulation (sim2sim) and simulation to reality (sim2real). The sim2sim domain adaptation setup is relatively similar to DeepMind Lab i.e. the source and target domains differ in terms of processes $\mathcal{G}_S$ and $\mathcal{G}_T$. However, there is a significant point of difference. In DeepMind Lab, all values of factors in the target domain, $\hat{s}_T$, are previously seen in the source domain; however, $\hat{s}_S \neq \hat{s}_T$ as the \emph{conjunctions} of these factor values are different. In sim2sim, by contrast, novel factor \emph{values} are experienced in the target domain (this accordingly also leads to novel factor conjunctions). Hence, DeepMind Lab may be considered to be assessing domain \emph{interpolation} performance, whereas sim2sim tests domain \emph{extrapolation}. 

The sim2real setup, on the other hand, is based on identical processes $\mathcal{G}_S = \mathcal{G}_T$, but different observation simulators $\bf{Sim}_S \neq \bf{Sim}_T$ corresponding to the MuJoCo simulation and the real world, which results in the so-called `perceptual reality gap' \citep{Rusu_etal_2016b}. More details of the tasks are given below.

\begin{figure}[t!]
\vskip 0.2in
\begin{center}
\centerline{\includegraphics[width=\columnwidth]{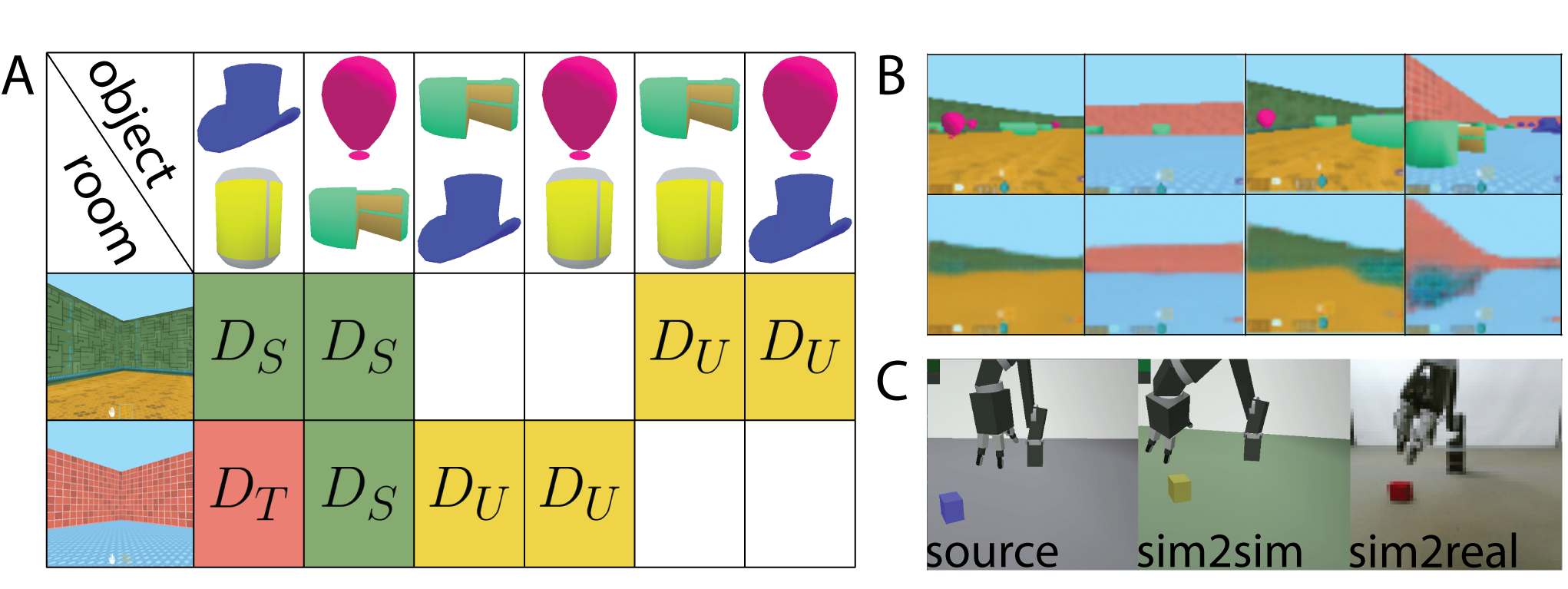}}
\vspace{-15pt}
\caption{\textbf{A}: DeepMind Lab \citep{deepmind_lab} transfer task setup. Different conjunctions of \{room, object$_1$, object$_2$\} were used during different parts of the domain adaptation curriculum.
During stage one, $D_U$ (shown in yellow), we used a minimal set spanning all objects and all rooms whereby each object is seen in each room. Note there is no extrinsic reward signal or notion of `task' in this phase.
During stage two, $D_S$ (shown in green), the RL agents were taught to pick up cans and balloons and avoid hats and cakes. The objects were always presented in pairs hat/can and cake/balloon. The agent never saw the hat/can pair in the pink room.
This novel room/object conjunction was presented as the target domain adaptation condition $D_T$ (shown in red) where the ability of the agent to transfer knowledge of the objects' value to a novel environment was tested. \textbf{B}: \betavae reconstructions (bottom row) using frames from DeepMind Lab (top row). Due to the increased $\beta>1$ necessary to disentangle the data generative factors of variations the model lost information about objects. See Fig.~\ref{fig_vae_ae_traversals} for a model appropriately capturing objects. \textbf{C}: \emph{left} -- sample frames from MuJoCo simulation environments used for vision (phase 1, $S_U$) and source policy training (phase 2, $S_S$); \emph{middle} -- sim2sim domain adaptation test (phase 3, $S_T$); and \emph{right} -- sim2real domain adaptation test (phase 3, $S_T$).}
\vspace{-30pt}
\label{fig_framework}
\end{center}
\end{figure}

\subsection{DeepMind Lab}
\label{sec_tasks_lab}
DeepMind Lab is a first person 3D game environment with rich visuals and realistic physics. We used a standard seek-avoid object gathering setup, where a room is initialised with an equal number of randomly placed objects of two different types. One of the object varieties is `good' (its collection is rewarded +1), while the other is `bad' (its collection is punished -1). The full state space $\hat{\mathcal{S}}$ consisted of all conjunctions of two room types (pink and green based on the colour of the walls) and four object types (hat, can, cake and balloon) (see Fig.~\ref{fig_framework}A). The source domain $D_S$ contained environments with hats/cans presented in the green room, and balloons/cakes presented in either the green or the pink room. The target domain $D_T$ contained hats/cans presented in the pink room. In both domains cans and balloons were the rewarded objects.


\emph{1) Learn to see}: we used \betavaedae to learn the disentangled latent state representation $s^z$ that includes both the room and the object generative factors of variation within DeepMind Lab. We had to use the high-level feature space of a pre-trained DAE within the \betavaedae framework (see Section~\ref{sec_perceptual_similarity_loss}), instead of the pixel space of vanilla \betavae, because we found that objects failed to reconstruct when using the values of $\beta$ necessary to disentangle the generative factors of variation within DeepMind Lab (see Fig.~\ref{fig_framework}B).


\betavaedae was trained on observations $s^o_U$ collected by an RL agent with a simple wall-avoiding policy $\pi_U$ (otherwise the training data was dominated by close up images of walls). In order to enable the model to learn $\mathcal{F}(s^o_U) \approx \hat{\mathcal{S}}$, it is important to expose the agent to at least a minimal set of environments that span the range of values for each factor, and where no extraneous correlations are added between different factors\footnote{In our setup of DeepMind Lab domain adaptation task, the object and environment factors are supposed to be independent. In order to ensure that \betavaedae learns a factorised representation that reflects this ground truth independence, we present observations of all possible conjunctions of room and individual object types.}(see Fig.~\ref{fig_framework}A, yellow). See Section~\ref{sec_sup_ae} in Supplementary Materials for details of \betavaedae training.

\emph{2) Learn to act}: the agent was trained with the algorithms detailed in Section~\ref{sec_reinforcement_learning_algorithms} on a seek-avoid task using the source domain ($D_S$) conjunctions of object/room shown in Fig.~\ref{fig_framework}A (green). Pre-trained \betavaedae from stage one was used as the `vision' part of various RL algorithms \citep[DQN, A3C and Episodic Control:][]{Mnih_etal_2015, Mnih_etal_2016, Blundell_etal_2016} to learn a source policy $\pi_S$ that picks up balloons and avoids cakes in both the green and the pink rooms, and picks up cans and avoids hats in the green rooms. See Section~\ref{sec_sup_ae} in Supplementary Materials for more details of the various versions of DARLA we have tried, each based on a different base RL algorithm.

\emph{3) Transfer}: we tested the ability of DARLA to transfer the seek-avoid policy $\pi_S$ it had learnt on the source domain in stage two using the domain adaptation condition $D_T$ illustrated in Figure~\ref{fig_framework}A (red). The agent had to continue picking up cans and avoid hats in the pink room, even though these objects had only been seen in the green room during source policy training. The optimal policy $\pi_T$ is one that maintains the reward polarity from the source domain (cans are good and hats are bad). For further details, see Appendix~\ref{sup_lab_task_details}.
 
\subsection{Jaco Arm and MuJoCo}
\label{sec_tasks_jaco}
We used frames from an RGB camera facing a robotic Jaco arm, or a matching rendered camera view from a MuJoCo physics simulation environment \citep{Todorov_etal_2012} to investigate the performance of DARLA in two domain adaptation scenarios: 1) simulation to simulation (sim2sim), and 2) simulation to reality (sim2real). The sim2real setup is of particular importance, since the progress that deep RL has brought to control tasks in simulation \citep{Schulman_etal_2015, Mnih_etal_2016, Levine_Abbeel_2014, Heess_etal_2015, Lillicrap_etal_2015, Schulman_etal_2016} has not yet translated as well to reality, despite various attempts \citep{Tobin_etal_2017, Tzeng_etal_2016, Daftry_etal_2016, Finn_etal_2015, Rusu_etal_2016b}. Solving control problems in reality is hard due to sparse reward signals, expensive data acquisition and the attendant danger of breaking the robot (or its human minders) during exploration.

In both sim2sim and sim2real, we trained the agent to perform an object reaching policy where the goal is to place the end effector as close to the object as possible. While conceptually the reaching task is simple, it is a hard control problem since it requires correct inference of the arm and object positions and velocities from raw visual inputs.

\emph{1) Learn to see}: \betavae was trained on observations collected in MuJoCo simulations with the same factors of variation as in $D_S$. In order to enable the model to learn $\mathcal{F}(s^o_U) \approx \hat{s}$, a reaching policy was applied to phantom objects placed in random positions - therefore ensuring that the agent learnt the independent nature of the arm position and object position (see Fig.~\ref{fig_framework}C, left);  

\emph{2) Learn to act}: a feedforward-A3C based agent with the vision module pre-trained in stage one was taught a source reaching policy $\pi_S$ towards the real object in simulation (see Fig.~\ref{fig_framework}C (left) for an example frame, and Sec.~\ref{sec_reinforcement_learning_algorithm_details} in Supplementary Materials for a fuller description of the agent). In the source domain $D_S$ the agent was trained on a distribution of camera angles and positions. The colour of the tabletop on which the arm rests and the object colour were both sampled anew every episode.

\emph{3) Transfer}: \textbf{sim2sim}: in the target domain, $D_T$, the agent was faced with a new distribution of camera angles and positions with little overlap with the source domain distributions, as well as a completely held out set of object colours (see Fig.~\ref{fig_framework}C, middle). \textbf{sim2real}: in the target domain $D_T$ the camera position and angle as well as the tabletop colour and object colour were sampled from the same distributions as seen in the source domain $D_S$, but the target domain $D_T$ was now the real world. Many details present in the real world such as shadows, specularity, multiple light sources and so on are not modelled in the simulation; the physics engine is also not a perfect model of reality. Thus sim2real tests the ability of the agent to cross the perceptual-reality gap and generalise its source policy $\pi_S$ to the real world (see Fig.~\ref{fig_framework}C, right). For further details, see Appendix~\ref{sup_jaco_task_details}.


\section{Results}
\label{sec_results}

\begin{figure}[t!]
\begin{center}
\centerline{\includegraphics[width=\columnwidth]{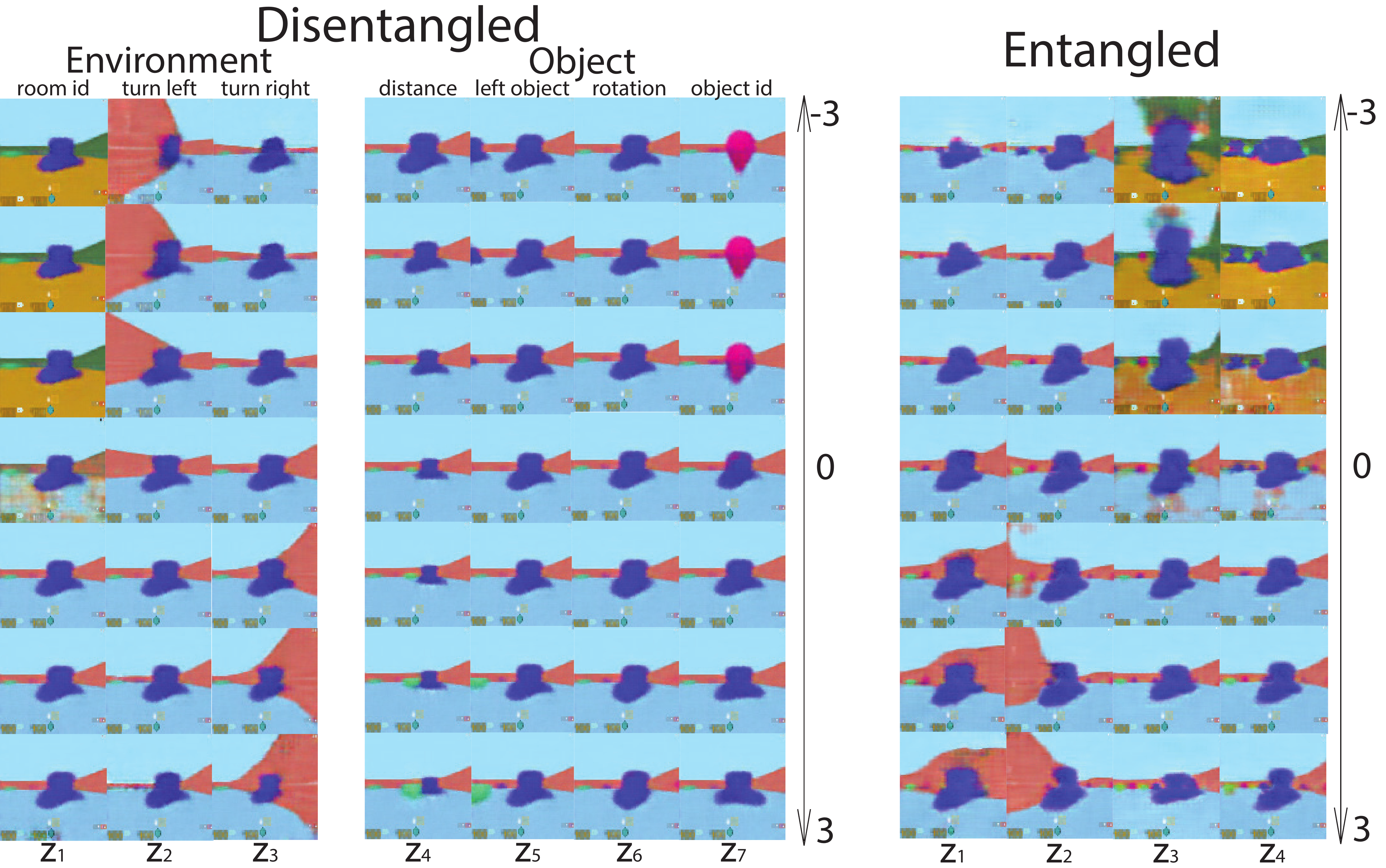}}
\vspace{-10pt}
\caption{Plot of traversals of various latents of an entangled and a disentangled version of \betavaedae using frames from DeepMind Lab \cite{deepmind_lab}.}
\vspace{-35pt}
\label{fig_vae_ae_traversals}
\end{center}
\end{figure}

\begin{figure}[t]
\vskip 0.2in
\begin{center}
\centerline{\includegraphics[width=\columnwidth]{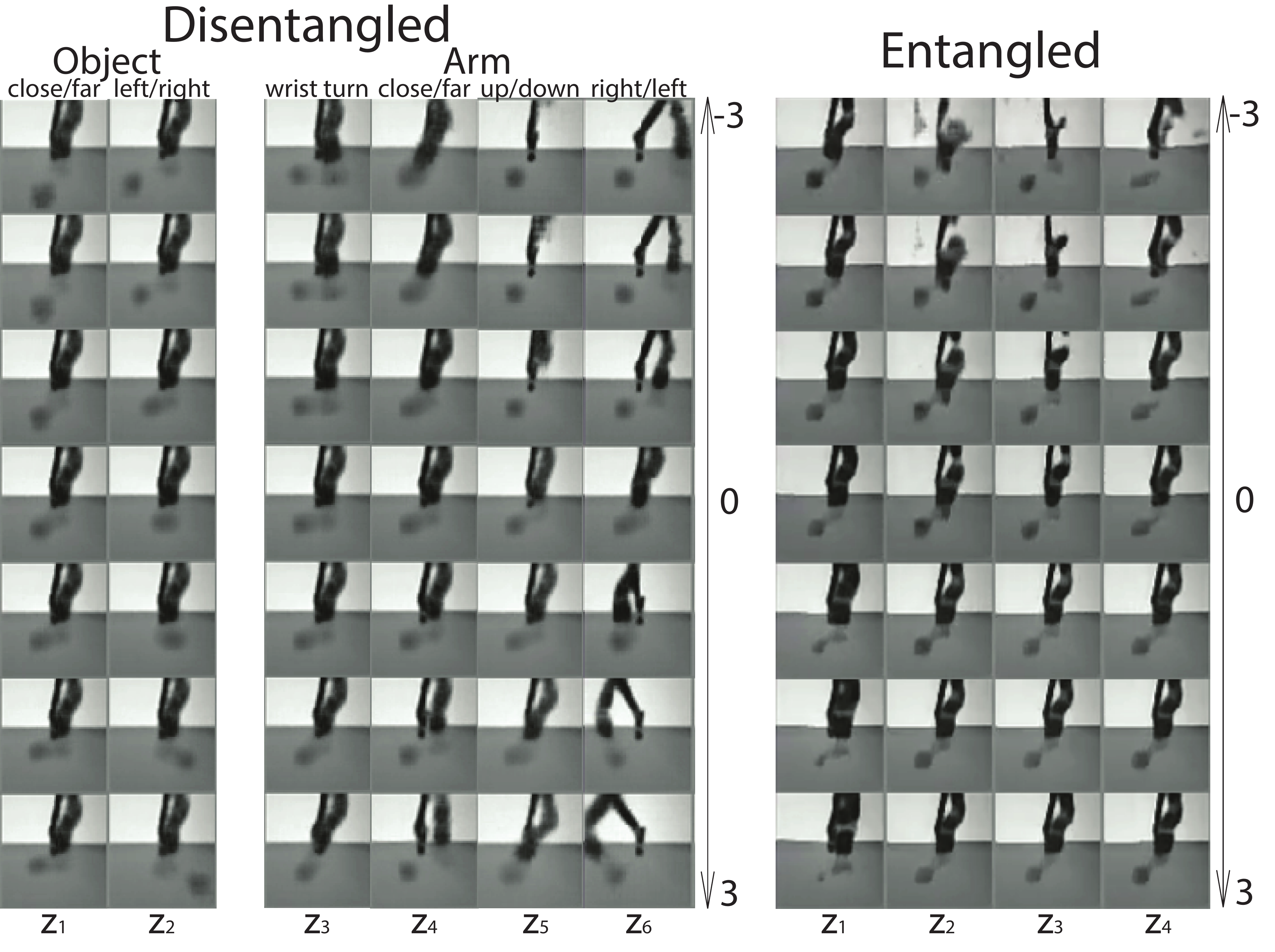}}
\vspace{-10pt}
\caption{Plot of traversals of \betavae on MuJoCo. Using a disentangled \betavae model, single latents directly control for the factors responsible for the object or arm placements.}
\vspace{-30pt}
\label{fig_mujoco_traversals}
\end{center}
\end{figure}

\begin{figure*}[ht]
  \includegraphics[width=\textwidth]{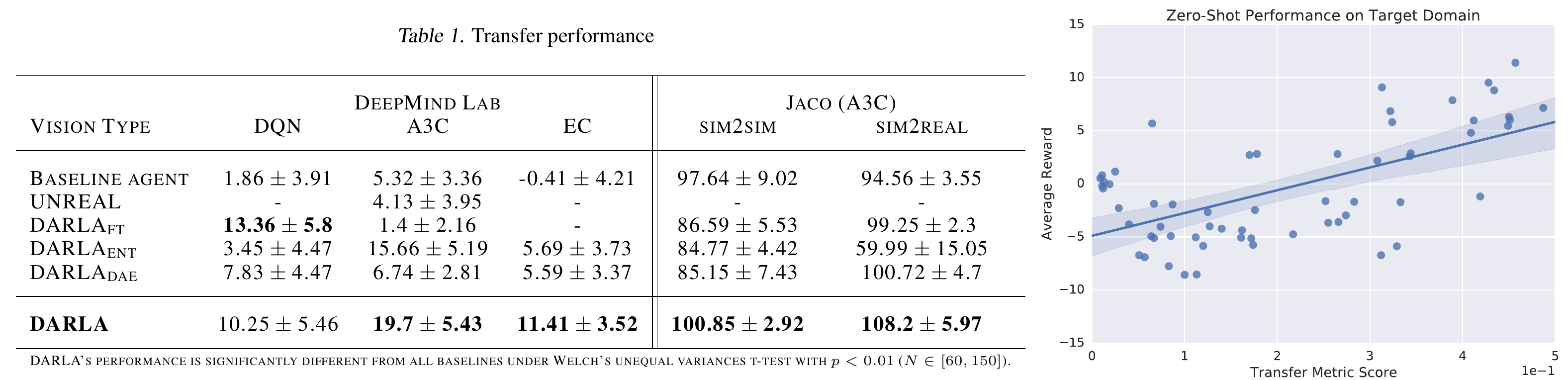}
  \vspace{-20pt}
  \caption{\textbf{Table}: Zero-shot performance (avg. reward per episode) of the source policy $\pi_S$ in target domains within DeepMind Lab and Jaco/MuJoCo environments. Baseline agent refers to vanilla DQN/A3C/EC (DeepMind Lab) or A3C (Jaco) agents. See main text for more detailed model descriptions. \textbf{Figure}: Correlation between zero-shot performance transfer performance on the DeepMind Lab task obtained by EC based DARLA and the level of disentanglement as measured by the transfer/disentanglement score ($r=0.6$, $p<0.001$)}    \label{fig_transfer_vs_disentanglement}
  \vspace{-10pt}
\end{figure*}

We evaluated the robustness of DARLA's policy $\pi_S$ learnt on the source domain to various shifts in the input data distribution. In particular, we used domain adaptation scenarios based on the DeepMind Lab seek-avoid task and the Jaco arm reaching task described in Sec.~\ref{sec_tasks}. On each task we compared DARLA's performance to that of various baselines. We evaluated the importance of learning `good' vision during stage one of the pipeline, i.e one that maps the input observations $s^o$ to disentangled representations $s^z \approx \hat{s}$. In order to do this, we ran the DARLA pipeline with different vision models: the encoders of a disentangled \betavae \kern-0.5ex \footnote{In this section of the paper, we use the term \betavae to refer to a standard \betavae for the MuJoCo experiments, and a \betavaedae for the DeepMind Lab experiments (as described in stage 1 of Sec.~\ref{sec_tasks_lab}).} (the original DARLA), an \emph{entangled} \betavae (DARLA$_\text{ENT}$), and a denoising autoencoder (DARLA$_\text{DAE}$). Apart from the nature of the learnt representations $s^z$, DARLA and all versions of its baselines were equivalent throughout the three stages of our proposed pipeline in terms of architecture and the observed data distribution (see Sec.~\ref{sec_model_details} in Supplementary Materials for more details).

Figs.~\ref{fig_vae_ae_traversals}-\ref{fig_mujoco_traversals} display the degree of disentanglement learnt by the vision modules of DARLA and DARLA$_\text{ENT}$ on DeepMind Lab and MuJoCo. DARLA's vision learnt to independently represent environment variables (such as room colour-scheme and geometry) and object-related variables (change of object type, size, rotation) on DeepMind Lab (Fig.~\ref{fig_vae_ae_traversals}, left). Disentangling was also evident in MuJoCo. Fig.~\ref{fig_mujoco_traversals}, left, shows that DARLA's single latent units $z_i$ learnt to represent different aspects of the Jaco arm, the object, and the camera. By contrast, in the representations learnt by DARLA$_\text{ENT}$, each latent is responsible for changes to both the environment and objects (Fig.~\ref{fig_vae_ae_traversals}, right) in DeepMind Lab, or a mixture of camera, object and/or arm movements (Fig.~\ref{fig_mujoco_traversals}, right) in MuJoCo.

The table in Fig.~\ref{fig_transfer_vs_disentanglement} shows the average performance (across different seeds) in terms of rewards per episode of the various agents on the target domain with no fine-tuning of the source policy $\pi_S$. It can be seen that DARLA is able to zero-shot-generalise significantly better than DARLA$_\text{ENT}$ or DARLA$_\text{DAE}$, highlighting the importance of learning a disentangled representation $s^z = s^z_{\hat{\mathcal{S}}}$ during the unsupervised stage one of the DARLA pipeline. In particular, this also demonstrates that the improved domain transfer performance is not simply a function of increased exposure to training observations, as both DARLA$_\text{ENT}$ and DARLA$_\text{DAE}$ were exposed to the same data. The results are mostly consistent across target domains and in most cases DARLA is significantly better than the second-best-performing agent. This holds in the sim2real task \footnote{See https://youtu.be/sZqrWFl0wQ4 for example sim2sim and sim2real zero-shot transfer policies of DARLA and baseline A3C agent.}, where being able to perform zero-shot policy transfer is highly valuable due to the particular difficulties of gathering data in the real world.

DARLA's performance is particularly surprising as it actually preserves \emph{less} information about the raw observations $s^o$ than DARLA$_\text{ENT}$ and DARLA$_\text{DAE}$. This is due to the nature of the \betavae and how it achieves disentangling; the disentangled model utilised a significantly higher value of the hyperparameter $\beta$ than the entangled model (see Appendix~\ref{sec_model_details} for further details), which constrains the capacity of the latent channel. Indeed, DARLA's \betavae only utilises 8 of its possible 32 Gaussian latents to store observation-specific information for MuJoCo/Jaco (and 20 in DeepMind Lab), whereas DARLA$_\text{ENT}$ utilises all 32 for both environments (as does DARLA$_\text{DAE}$).

Furthermore, we examined what happens if DARLA's vision (i.e. the encoder of the disentangled \betavae \kern-0.5ex) is allowed to be fine-tuned via gradient updates while learning the source policy during stage two of the pipeline. This is denoted by DARLA$_\text{FT}$ in the table in Fig.~\ref{fig_transfer_vs_disentanglement}. We see that it exhibits significantly worse performance than that of DARLA in zero-shot domain adaptation using an A3C-based agent in all tasks. This suggests that a favourable initialisation does not make up for subsequent overfitting to the source domain for the on-policy A3C. However, the off-policy DQN-based fine-tuned agent performs very well. We leave further investigation of this curious effect for future work.

Finally, we compared the performance of DARLA to an UNREAL \citep{Jaderberg_etal_2017} agent with the same architecture. Despite also exploiting the unsupervised data available in the source domain, UNREAL performed worse than baseline A3C on the DeepMind Lab domain adaptation task. This further demonstrates that use of unsupervised data in itself is not a panacea for transfer performance; it must be utilised in a careful and structured manner conducive to learning disentangled latent states $s^z = s^z_{\hat{\mathcal{S}}}$.

In order to quantitatively evaluate our hypothesis that disentangled representations are essential for DARLA's performance in domain adaptation scenarios, we trained various DARLAs with different degrees of learnt disentanglement in $s^z$ by varying $\beta$ (of \betavae \kern-0.5ex) during stage one of the pipeline. We then calculated the correlation between the performance of the EC-based DARLA on the DeepMind Lab domain adaptation task and the transfer metric, which approximately measures the quality of disentanglement of DARLA's latent representations $s^z$ (see Sec.~\ref{sec_transfer_metric_details} in Supplementary Materials). This is shown in the chart in Fig.~\ref{fig_transfer_vs_disentanglement}; as can be seen, there is a strong positive correlation between the level of disentanglement and DARLA's zero-shot domain transfer performance ($r=0.6$, $p<0.001$).


Having shown the robust utility of disentangled representations in agents for domain adaptation, we note that there is evidence that they can provide an important additional benefit. We found significantly improved speed of learning of $\pi_S$ on the source domain itself, as a function of how disentangled the model was. The gain in data efficiency from disentangled representations for source policy learning is not the main focus of this paper so we leave it out of the main text; however, we provide results and discussion in Section~\ref{sec_sup_source_task_perf} in Supplementary Materials.

\section{Conclusion}
We have demonstrated the benefits of using disentangled representations in a deep RL setting for domain adaptation. In particular, we have proposed DARLA, a multi-stage RL agent. DARLA first learns a visual system that encodes the observations it receives from the environment as disentangled representations, in a completely unsupervised manner. It then uses these representations to learn a robust source policy that is capable of zero-shot domain adaptation. 

We have demonstrated the efficacy of this approach in a range of domains and task setups: a 3D naturalistic first-person environment (DeepMind Lab), a simulated graphics and physics engine (MuJoCo), and crossing the simulation to reality gap (MuJoCo to Jaco sim2real). We have also shown that the effect of disentangling is consistent across very different RL algorithms (DQN, A3C, EC), achieving significant improvements over the baseline algorithms (median 2.7 times improvement in zero-shot transfer across tasks and algorithms). To the best of our knowledge, this is the first comprehensive empirical demonstration of the strength of disentangled representations for domain adaptation in a deep RL setting.

\newpage
\small
\bibliography{bibliography}
\bibliographystyle{icml2017}


\newpage

\appendix
\section{Supplementary Materials}

\subsection{The Reinforcement Learning Paradigm}
\label{sec_reinforcement_learning_appendix}
The reinforcement learning (RL) paradigm consists of an agent receiving a sequence of observations $s^o_t$ which are some function of environment states $s_t \in \mathcal{S}$ and may be accompanied by rewards $r_{t+1} \in R$ conditional on the actions $a_t \in \mathcal{A}$, chosen at each time step $t$ \citep{Sutton_Barto_1998}. We assume that these interactions can be modelled as a Markov Decision Process (MDP) \citep{Puterman_1994} defined as a tuple $D \equiv (\mathcal{S}, \mathcal{A}, \mathcal{T}, R, \gamma )$.  $\mathcal{T} = p(s | s_t, a_t)$ is a transition function that models the distribution of all possible next states given action $a_t$ is taken in state $s_t$ for all $s_t \in \mathcal{S}$ and $a_t \in \mathcal{A}$. Each transition $s_t \overset{a_t}{\rightarrow} s_{t+1}$ may be accompanied by a reward signal $r_{t+1}(s_t, a_t, s_{t+1})$. The goal of the agent is to learn a policy $\pi(a_t|s_t)$, a probability distribution over actions $a_t \in  \mathcal{A}$, that maximises the expected return i.e. the discounted sum of future rewards $R_{t} = \mathbb{E}[\sum_{\tau=1}^{T-t}\gamma^{\tau-1}r_{t+\tau}]$. $T$ is the time step at which each episode ends, and  $\gamma \in [0, 1)$ is the discount factor that progressively down-weights future rewards. Given a policy $\pi(a|s)$, one can define the value function $V_{\pi}(s) = \mathbb{E} [R_t|s_t=s, \pi]$, which is the expected return from state $s$ following policy $\pi$. The action-value function $Q_{\pi}(s, a) = \mathbb{E} [R_t|s_t=s, a_t=a, \pi]$ is the expected return for taking action $a$ in state $s$ at time $t$, and then following policy $\pi$ from time $t+1$ onward.

\subsection{Further task details}
\label{sup_task_details}

\subsubsection{DeepMind Lab}
\label{sup_lab_task_details}
As described in Sec~\ref{sec_tasks_lab}, in each source episode of DeepMind Lab the agent was presented with one of three possible room/object type conjunctions, chosen at random. These are marked $D_S$ in Fig~\ref{fig_framework}. The setup was a seek-avoid style task, where one of the two object types in the room gave a reward of +1 and the other gave a reward of -1. The agent was allowed to pick up objects for 60 seconds after which the episode would terminate and a new one would begin; if the agent was able to pick up all the `good' objects in less than 60 seconds, a new episode was begun immediately. The agent was spawned in a random location in the room at the start of each new episode.

During transfer, the agent was placed into the held out conjunction of object types and room background; see $D_T$ in Fig~\ref{fig_framework}.

Visual pre-training was performed in other conjunctions of object type and room background denoted $D_U$ in Fig~\ref{fig_framework}.

The observation size of frames in the DeepMind Lab task was 84x84x3 ($H$x$W$x$C$).

\subsubsection{MuJoCo/Jaco Arm Experiments}
\label{sup_jaco_task_details}
As described in Sec~\ref{sec_tasks_jaco}, the source task consisted of an agent learning to control a simulated arm in order to reach toward an object. A shaping reward was used, with a maximum value of 1 when the centre of the object fell between the pinch and grip sites of the end effector, or within a 10cm distance of the two. Distances on the x and y dimensions counted double compared to distances on the z dimension.

During each episode the object was placed at a random drop point within a 40x40cm area, and the arm was set to a random initial start position high above the work-space, independent of the object's position. Each episode lasted for 150 steps, or 7.5 seconds, with a control step of 50ms. Observations $s^o_U$ were sampled randomly across episodes. Overall, 4 million frames of dimensions 64x64x3 ($H$x$W$x$C$) were used for this stage of the curriculum. For each episode the camera position and orientation were randomly sampled from an isotropic normal distribution centred around the approximate position and orientation of the real camera, with standard deviation 0.01. No precise measurements were used to match the two. Work-space table colour was sampled uniformly between $-5\%$ and $+5\%$ around the midpoint, independently for each RGB channel; object colours were sampled uniformly at random in RGB space, rejecting colours which fell within a ball around 10 held-out intensities (radius $10\%$ of range); the latter were only used for simulated transfer experiments, i.e. in $D_T$ in the sim2sim experiments. Additionally, Gaussian noise with standard deviation 0.01 was added to the observations $s^o_T$ in the sim2sim task.

For the real Jaco arm and its MuJoCo simulation counterpart, each of the nine joints could independently take 11 different actions (a linear discretisation of the continuous velocity action space).  In simulation Gaussian noise with standard deviation 0.1 was added to each discrete velocity output; delays in the real setup between observations and action execution were simulated by randomly mixing velocity outputs from two previous steps instead of emitting the last output directly. Speed ranges were between $-50\%$ and $50\%$ of the Jaco arm's top speed on joints 1 through 6 starting at the base, while the fingers could use a full range. For safety reasons the speed ranges have been reduced by a factor of 0.3 while evaluating agents on the Jaco arm, without significant performance degradation.

\subsection{Vision model details}
\label{sec_model_details}

\subsubsection{Denoising Autoencoder for \betavae}
\label{sec_sup_ae}

A denoising autoencoder (DAE) was used as a model to provide the feature space for the \betavae reconstruction loss to be computed over (for motivation, see Sec.~\ref{sec_perceptual_similarity_loss}). The DAE was trained with occlusion-style masking noise in the vein of \citep{Pathak_etal_2016}, with the aim for the DAE to learn a semantic representation of the input frames. Concretely, two values were independently sampled from $U[0, W]$ and two from $U[0, H]$ where $W$ and $H$ were the width and height of the input frames. These four values determined the corners of the rectangular mask applied; all pixels that fell within the mask were set to zero.

The DAE architecture consisted of four convolutional layers, each with kernel size 4 and stride 2 in both the height and width dimensions. The number of filters learnt for each layer was \{32, 32, 64, 64\} respectively. The bottleneck layer consisted of a fully connected layer of size 100 neurons. This was followed by four deconvolutional layers, again with kernel sizes 4, strides 2, and \{64, 64, 32, 32\} filters. The padding algorithm used was `SAME' in TensorFlow \citep{Abadi_etal_2015}. ReLU non-linearities were used throughout.

The model was trained with loss given by the L2 distance of the outputs from the original, un-noised inputs. The optimiser used was Adam \citep{Kingma_Ba_2014} with a learning rate of 1e-3.

\subsubsection{\betavae with Perceptual Similarity Loss}
\label{sec_sup_betavae_with_perceptual_similarity_loss}

After training a DAE, as detailed in the previous section\footnote{In principle, the \betavaedae could also have been trained end-to-end in one pass, but we did not experiment with this.}, a \betavaedae was trained with perceptual similarity loss given by Eq.~\ref{eq_beta_vae_ae}, repeated here:
\begin{align}
\mathcal{L}(\theta, \phi; \v{x}, \v{z}, \beta)  =& \mathbb{E}_{q_\phi(\v{z}|\v{x})}\left \|J(\hat{\v{x}}) - J(\v{x})  \right \|_2^2   \nonumber \\
 &- \beta \ D_{KL}(q_\phi(\v{z}|\v{x}) || p(\v{z})) \label{eq_beta_vae_ae_2}
\end{align}
Specifically, the input was passed through the \betavae and a sampled\footnote{It is more typical to use the mean of the reconstruction distribution, but this does not induce any pressure on the Gaussians parametrising the decoder to reduce their variances. Hence full samples were used instead.} reconstruction was passed through the pre-trained DAE up to a designated layer. The L2 distance of this representation from the representation of the original input passed through the same layers of the DAE was then computed, and this formed the training loss for the \betavae part of the \betavaedae\footnote{The representations were taken after passing through the layer but before passing through the following non-linearity. We also briefly experimented with taking the L2 loss post-activation but did not find a significant difference.}. The DAE weights remained frozen throughout.

The \betavae architecture consisted of an encoder of four convolutional layers, each with kernel size 4, and stride 2 in the height and width dimensions. The number of filters learnt for each layer was \{32, 32, 64, 64\} respectively. This was followed by a fully connected layer of size 256 neurons. The latent layer comprised 64 neurons parametrising 32 (marginally) independent Gaussian distributions. The decoder architecture was simply the reverse of the encoder, utilising deconvolutional layers. The decoder used was Gaussian, so that the number of output channels was $2C$, where $C$ was the number of channels that the input frames had. The padding algorithm used was `SAME' in TensorFlow. ReLU non-linearities were used throughout.

The model was trained with the loss given by Eq.~\ref{eq_beta_vae_ae_2}. Specifically, the disentangled model used for DARLA was trained with a $\beta$ hyperparameter value of 1 and the layer of the DAE used to compute the perceptual similarity loss was the last deconvolutional layer. The entangled model used for DARLA$_\text{ENT}$ was trained with a $\beta$ hyperparameter value of 0.1 with the last deconvolutional layer of the DAE was used to compute the perceptual similarity loss.

The optimiser used was Adam with a learning rate of 1e-4.

\subsubsection{\betavae}
\label{sup_betavae_details}
For the MuJoCo/Jaco tasks, a standard \betavae was used rather than the \betavaedae used for DeepMind Lab. The architecture of the VAE encoder, decoder and the latent size were exactly as described in the previous section~\ref{sec_sup_betavae_with_perceptual_similarity_loss}. $\beta$ for the the disentangled \betavae in DARLA was 175. $\beta$ for the entangled model DARLA$_\text{ENT}$ was 1, corresponding to the standard VAE of \citep{Kingma_Welling_2014}.

The optimizer used was Adam with a learning rate of 1e-4.

\subsubsection{Denoising Autoencoder for baseline}
For the baseline model DARLA$_\text{DAE}$, we trained a denoising autoencoder with occlusion-style masking noise as described in Appendix Section~\ref{sec_sup_ae}. The architecture used matched that exactly of the \betavae described in Appendix Section~\ref{sec_sup_betavae_with_perceptual_similarity_loss} - however, all stochastic nodes were replaced with deterministic neurons.

The optimizer used was Adam with a learning rate of 1e-4.

\subsection{Reinforcement Learning Algorithm Details}
\label{sec_reinforcement_learning_algorithm_details}

\subsubsection{DeepMind Lab}
The action space in the DeepMind Lab task consisted of 8 discrete actions.

\textbf{DQN:} in DQN, the convolutional (or `vision') part of the Q-net was replaced with the encoder of the \betavaedae from stage 1 and frozen. DQN takes four consecutive frames as input in order to capture some aspect of environment dynamics in the agent's state. In order to match this in our setup with a pre-trained vision stack $\mathcal{F}_U$, we passed each observation frame $s^o_{\{1..4\}}$ through the pre-trained model $s^z_{\{1..4\}} = \mathcal{F}_U(s^o_{\{1..4\}})$  and then concatenated the outputs together to form the k-dimensional (where $k = 4|s^z|$) input to the policy network. In this case the size of $s^z$ was 64 for DARLA as well as DARLA$_\text{ENT}$, DARLA$_\text{DAE}$ and DARLA$_\text{FT}$.

On top of the frozen convolutional stack, two `policy' layers of 512 neurons each were used, with a final linear layer of 8 neurons corresponding to the size of the action space in the DeepMind Lab task. ReLU non-linearities were used throughout. All other hyperparameters were as reported in \citep{Mnih_etal_2015}.

\textbf{A3C:} in A3C, as with DQN, the convolutional part of the network that is shared between the policy net and the value net was replaced with the encoder of the \betavaedae in DeepMind Lab tasks. All other hyperparameters were as reported in \citep{Mnih_etal_2016}.

\textbf{Episodic Control:} for the Episodic Controller-based DARLA we used mostly the same hyperparameters as in the original paper by \citep{Blundell_etal_2016}. We explored the following hyperparameter settings: number of nearest neighbours $\in {\{10, 50\}}$, return horizon $\in \{100, 400, 800, 1800, 500000\}$, kernel type $\in$ \{inverse, gaussian\}, kernel width $\in \{1e-6, 1e-5, 1e-4, 1e-3, 1e-2, 1e-1, 0.5, 0.99\}$ and we tried training EC with and without Peng's $Q(\lambda)$ \citep{Peng_1993}. In practice we found that none of the explored hyperparameter choices significantly influenced the results of our experiments. The final hyperparameters used for all experiments reported in the paper were the following: number of nearest neighbours: 10, return horizon: 400, kernel type: inverse, kernel width: 1e-6 and no Peng's $Q(\lambda)$ \citep{Peng_1993}.

\textbf{UNREAL:} We used a vanilla version of UNREAL, with parameters as reported in \citep{Jaderberg_etal_2017}.

\subsubsection{MuJoCo/Jaco Arm Experiments}

For the real Jaco arm and its MuJoCo simulation, each of the nine joints could independently take 11 different actions (a linear discretisation of the continuous velocity action space). Therefore the action space size was 99.

DARLA for MuJoCo/Jaco was based on feedforward A3C \citep{Mnih_etal_2016}. We closely followed the simulation training setup of \citep{Rusu_etal_2016b} for feed-forward networks using raw visual-input only. In place of the usual conv-stack, however, we used the encoder of the \betavae as described in Appendix \ref{sup_betavae_details}. This was followed by a linear layer with 512 units, a ReLU non-linearity and a collection of 9 linear and softmax layers for the 9 independent policy outputs, as well as a single value output layer that outputted the value function.

\subsection{Disentanglement Evaluation}
\subsubsection{Visual Heuristic Details}
\label{sec_visual_heuristic_details}
In order to choose the optimal value of $\beta$ for the \betavae-DAE models and evaluate the fitness of the representations $s^z_U$ learnt in stage 1 of our pipeline (in terms of disentanglement achieved), we used the visual inspection heuristic described in \cite{Higgins_et_al_2017}. The heuristic involved clustering trained \betavae \ based models based on the number of informative latents (estimated as the number of latents $z_i$ with average inferred standard deviation below 0.75). For each cluster we examined the degree of learnt disentanglement by running inference on a number of seed images, then traversing each latent unit $z_{\left \{ i \right \}}$ one at a time over three standard deviations away from its average inferred mean while keeping all other latents $z_{\left \{ \setminus i \right \}}$ fixed to their inferred values. This allowed us to visually examine whether each individual latent unit $z_i$ learnt to control a single interpretable factor of variation in the data. A similar heuristic has been the \textit{de rigueur} method for exhibiting disentanglement in the disentanglement literature \citep{Chen_etal_2016, Kulkarni_etal_2015}.

\begin{figure}[t!]
\vskip 0.2in
\begin{center}
\centerline{\includegraphics[width=0.5\columnwidth]{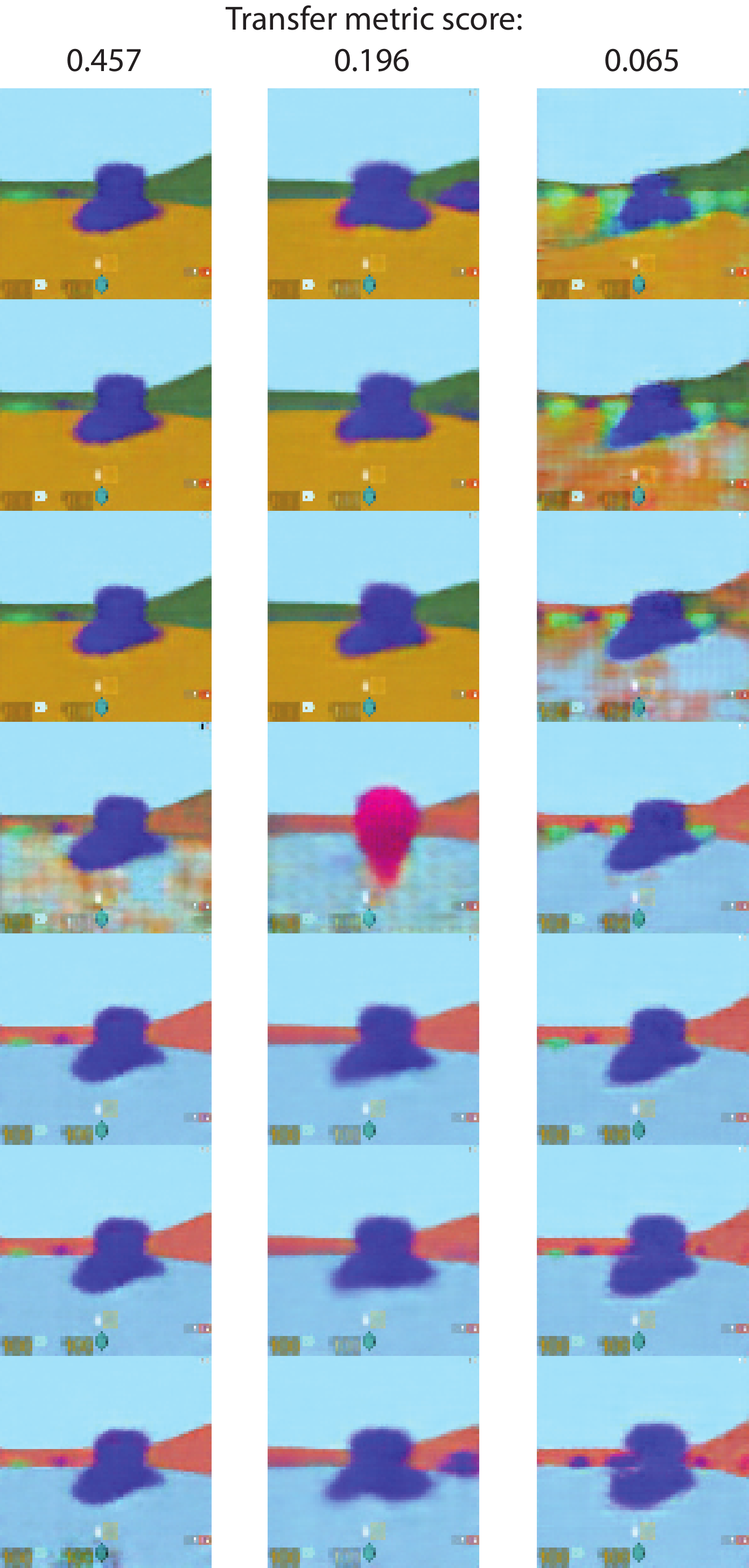}}
\caption{Traversals of the latent corresponding to room background for models with different transfer metric scores (shown top). Note that in the entangled model, many other objects appear and blue hat changes shape in addition to the background changing. For the model with middling transfer score, both the object type and background alter; whereas for the disentangled model, very little apart from the background changes. }
\label{fig_transfer_metric_vs_disentanglement}
\end{center}
\vskip -0.2in
\end{figure} 

\subsubsection{Transfer Metric Details}
\label{sec_transfer_metric_details}
In the case of DeepMind Lab, we were able to use the ground truth labels corresponding to the two factors of variation of the object type and the background to design a proxy to the disentanglement metric proposed in \citep{Higgins_et_al_2017}. The procedure used consisted of the following steps:
 
1) Train the model under consideration on observations $s^o_U$ to learn $\mathcal{F_U}$, as described in stage 1 of the DARLA pipeline.

2) Learn a linear model $\mathcal{L}: S^z_V \rightarrow M \times N$ from the representations $s^z_V = \mathcal{F}_V (s^o_V)$, where $M \in \{0, 1\}$ corresponds to the set of possible rooms and $N \in \{0, 1, 2, 3\}$ corresponds to the set of possible objects\footnote{For the purposes of this metric, we utilised rooms with only single objects, which we denote by the subscript $V$ e.g. the observation set $S^o_V$.}. Therefore we are learning a low-VC dimension classifier to predict the room and the object class from the latent representation of the model. Crucially, the linear model $\mathcal{L}$ is trained on only a subset of the Cartesian product $M \times N$ e.g. on $\{\{0, 0\}, \{0, 3\}, \{1, 1\}, \{1, 2\}\}$. In practice, we utilised a softmax classifier each for $M$ and $N$ and trained this using backpropagation with a cross-entropy loss, keeping the unsupervised model (and therefore $\mathcal{F_U}$) fixed.

3) The trained linear model $\mathcal{L}$'s accuracy is evaluated on the held out subset of the Cartesian product $M \times N$.

Although the above procedure only measures disentangling up to linearity, and only does so for the latents of object type and room background, we nevertheless found that the metric was highly correlated with disentanglement as determined via visual inspection (see Fig.~\ref{fig_transfer_metric_vs_disentanglement}).

\subsection{Background on RL Algorithms}
\label{sec_background_on_rl_algorithms}
In this Appendix, we provide background on the different RL algorithms that the DARLA framework was tested on in this paper.

\subsubsection{DQN}

(DQN) \citep{Mnih_etal_2015} is a variant of the Q-learning algorithm \citep{Watkins_1989} that utilises deep learning. It uses a neural network to parametrise an approximation for the action-value function $Q(s, a; \theta)$ using parameters $\theta$. These parameters are updated by minimising the mean-squared error of a 1-step lookahead loss $\mathcal{L}_Q = \mathbb{E}\left [ (r_t + \gamma max_{a'} Q(s', a'; \theta^-) - Q(s,a;\theta))^2 \right ]$, where $\theta^-$ are parameters corresponding to a frozen network and optimisation is performed with respect to $\theta$, with $\theta^-$ being synced to $\theta$ at regular intervals.

\subsubsection{A3C}

\emph{Asynchronous Advantage Actor-Critic}
(A3C) \citep{Mnih_etal_2016} is an asynchronous implementation of the advantage actor-critic paradigm \citep{Sutton_Barto_1998, Degris_etal_2012}, where separate threads run in parallel and perform updates to shared parameters. The different threads each hold their own instance of the environment and have different exploration policies, thereby decorrelating parameter updates without the need for experience replay.

A3C uses neural networks to approximate both policy $\pi(a|s; \theta)$ and value $V_{\pi}(s; \theta)$ functions using parameters $\theta$ using n-step look-ahead loss \citep{Peng_Williams_1996}. The algorithm is trained using an advantage actor-critic loss function with an entropy regularisation penalty: $\mathcal{L}_{A3C} \approx \mathcal{L}_{VR} + \mathcal{L}_{\pi} - \mathbb{E}_{s\sim \pi} \left [ \alpha H(\pi(a|s; \theta)) \right ]$, where $H$ is entropy. The parameter updates are performed after every $t_{max}$ actions or when a terminal state is reached. $\mathcal{L}_{VR}  = \mathbb{E}_{s\sim \pi}\left [ (R_{t:t+n} + \gamma ^n V(s_{t+n+1}; \theta) - V(s_{t};\theta))^2 \right ]$ and $\mathcal{L}_{\pi}  = \mathbb{E}_{s\sim \pi}\left [ log~ \pi(a|s; \theta)(Q^{\pi}(s, a; \theta) - V^{\pi}(s; \theta)) \right ]$. Unlike DQN, A3C uses an LSTM core to encode its history and therefore has a longer term memory permitting it to perform better in partially observed environments. In the version of A3C used in this paper for the DeepMind Lab task, the policy net additionally takes the last action $a_{t-1}$ and last reward $r_{t-1}$ as inputs along with the observation $s^o_t$, as introduced in \citep{Jaderberg_etal_2017}.

\subsubsection{UNREAL}

The \emph{UNREAL} agent \citep{Jaderberg_etal_2017} takes as a base an LSTM A3C agent \citep{Mnih_etal_2016} and augments it with a number of unsupervised auxiliary tasks that make use of the rich perceptual data available to the agent besides the (sometimes very sparse) extrinsic reward signals. This auxiliary learning tends to improve the representation learnt by the agent. While training the base agent, its observations, rewards, and actions are stored in a replay buffer, which is used by the auxiliary learning tasks. The tasks include: 1) pixel control – the agent learns how to control the environment by training auxiliary policies to maximally change pixel intensities in different parts of the input; 2) reward prediction - given a replay buffer of observations within a short time period of an extrinsic reward, the agent has to predict the reward obtained during the next unobserved timestep using a sequence of three preceding steps; 3) value function replay - extra training of the value function to promote faster value iteration.

\subsubsection{Episodic Control}

In its simplest form EC is a lookup table of states and actions denoted as $Q^{EC}(s, a)$. In each state EC picks the action with the highest $Q_{EC}$ value. At the end of each episode $Q^{EC}(s, a)$ is set to $R_t$ if $(s_t, a_t) \notin Q^{EC}$, where $R_t$ is the discounted return. Otherwise $Q^{EC}(s, a) = max\left \{ Q^{EC}(s,a), R_t \right \}$. In order to generalise its policy to novel states that are not in $Q^{EC}$, EC uses a non-parametric nearest neighbours search $\widehat{Q^{EC}}(s, a) = \frac{1}{k} \sum_{i=1}^k Q^{EC}(s^i, a)$, where $s^i, i=1,...,k$ are $k$ states with the smallest distance to the novel state $s$. Like DQN, EC takes a concatenation of four frames as input.

The EC algorithm is proposed as a model of fast hippocampal instance-based learning in the brain \citep{Marr_1971, Sutherland_Rudy_1989}, while the deep RL algorithms described above are more analogous to slow cortical learning that relies on generalised statistical summaries of the input distribution \citep{McClelland_etal_1995, Norman_OReilly_2003, Tulving_etal_1991}.

\subsection{Source Task Performance Results}
\label{sec_sup_source_task_perf}

\begin{figure}[h!]
\vskip 0.2in
\begin{center}
\centerline{\includegraphics[width=\columnwidth]{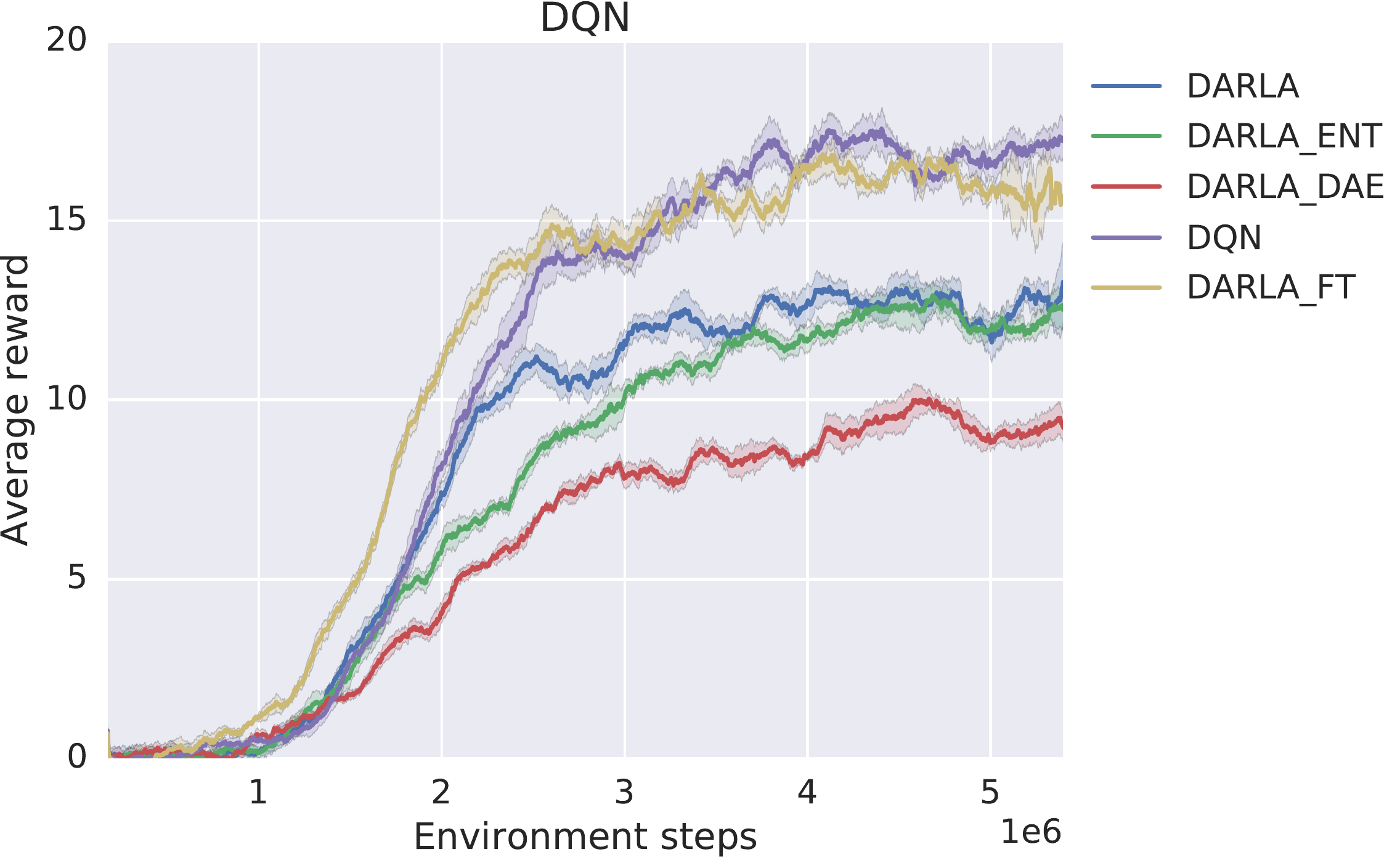}}
\vspace{-10pt}
\caption{Source task training curves for DQN. Curves show average and standard deviation over 20 random seeds.}
\label{fig_dqn_train}
\end{center}
\vskip -0.2in
\end{figure}

\begin{figure}[h!]
\vskip 0.2in
\begin{center}
\centerline{\includegraphics[width=0.9\columnwidth]{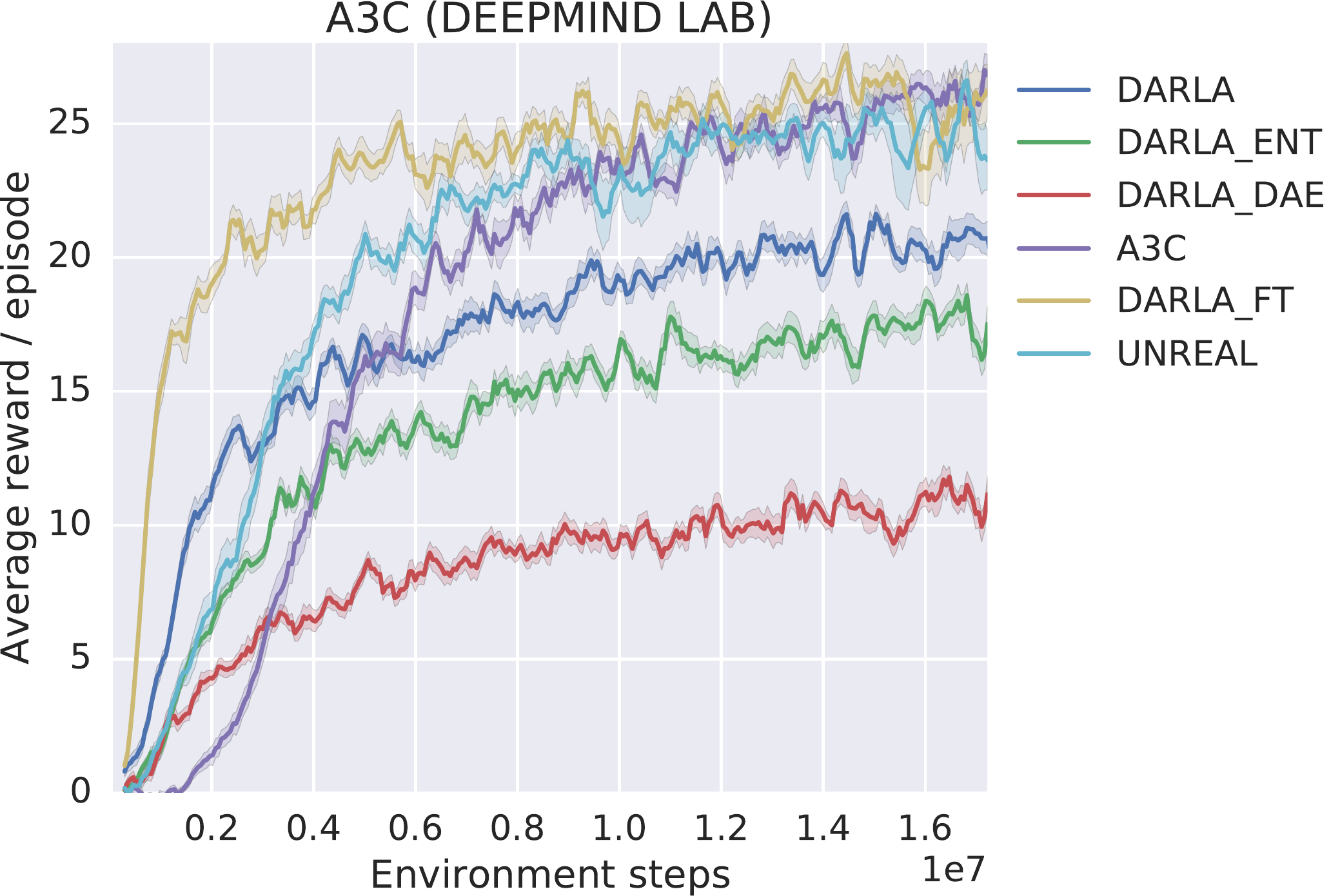}}
\vspace{-10pt}
\caption{Source task performance training curves for A3C and UNREAL. DARLA shows accelerated learning of the task compared to other architectures. Results show average and standard deviation over 20 random seeds, each using 16 workers.}
\label{fig_a3c_train}
\end{center}
\vskip -0.2in
\end{figure}

\begin{figure}[h!]
\vskip 0.2in
\begin{center}
\centerline{\includegraphics[width=0.9\columnwidth]{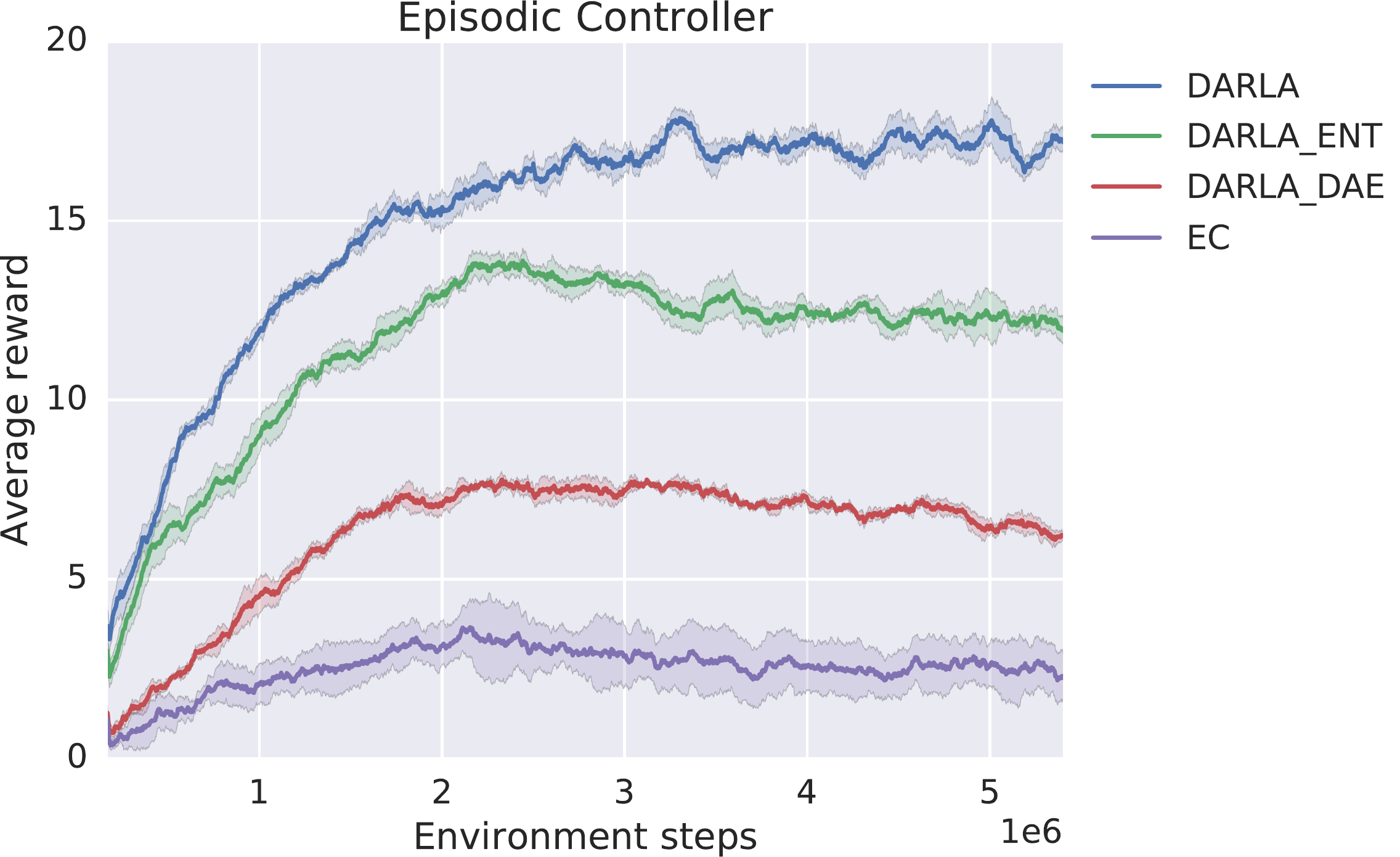}}
\vspace{-10pt}
\caption{Source task training curves for EC. Results show average and standard deviation over 20 random seeds.}
\label{fig_ec_train}
\end{center}
\vskip -0.2in
\end{figure}

\begin{figure}[h!]
\vskip 0.2in
\begin{center}
\centerline{\includegraphics[width=0.9\columnwidth]{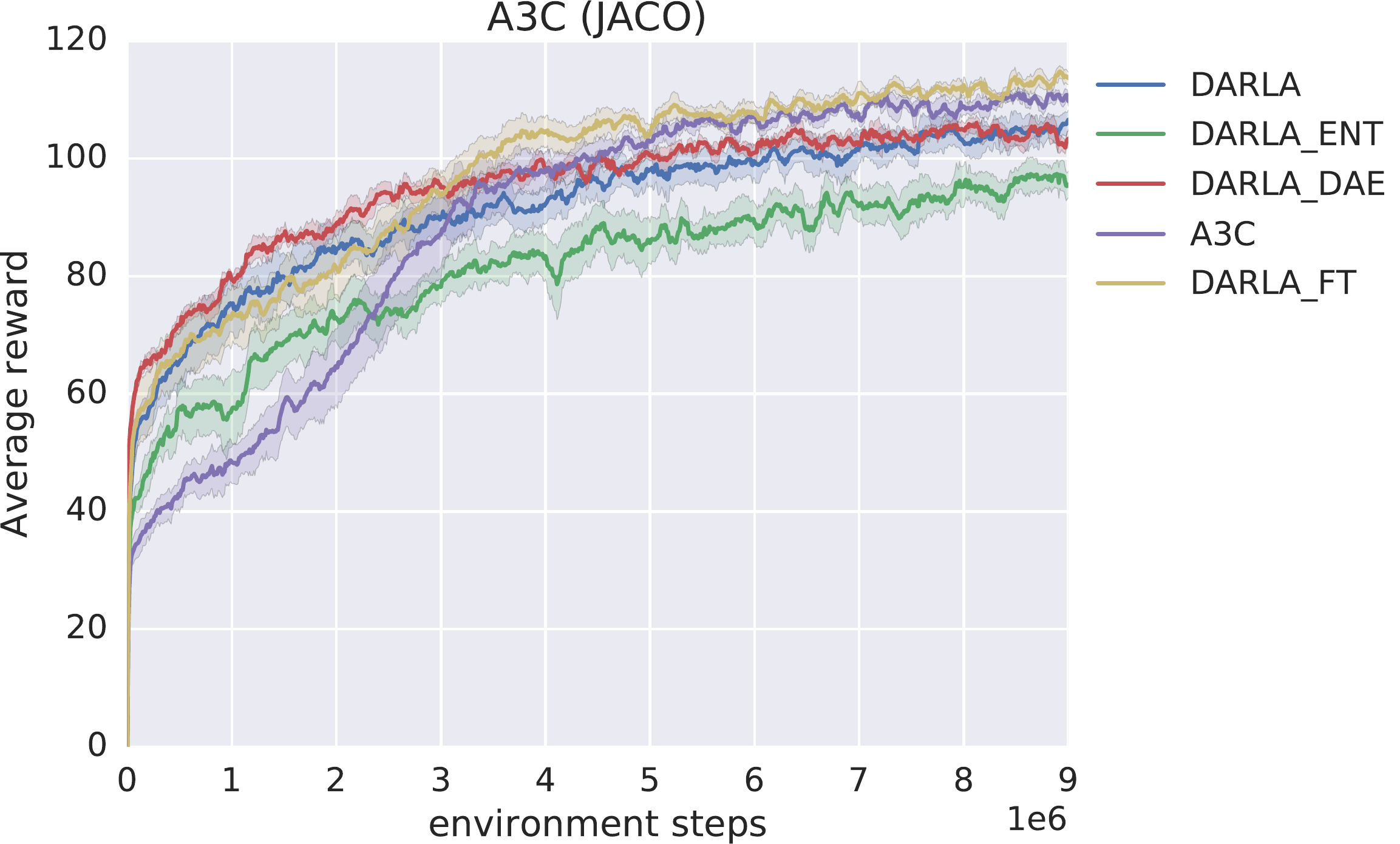}}
\caption{Training curves for various baselines on the source MuJoCo reaching task}
\label{fig_mujoco_a3c}
\end{center}
\vskip -0.2in
\end{figure}

The focus of this paper is primarily on zero-shot domain adaptation performance. However, it is also interesting to analyse the effect of the DARLA approach on source domain policy performance. In order to compare the models' behaviour on the source task, we examined the training curves (see Figures~\ref{fig_dqn_train}-\ref{fig_mujoco_a3c}) and noted in particular their:
\begin{enumerate}
    \item Asymptotic task performance, i.e. the rewards per episode at the point where $\pi_S$ has converged for the agent under consideration.
    \item Data efficiency, i.e. how quickly the training curve was able to achieve convergence.
\end{enumerate}

We note the following consistent trends across the results:
\begin{enumerate}
    \item Using DARLA provided an initial boost in learning performance, which depended on the degree of disentanglement of the representation. This was particularly observable in A3C, see Fig.~\ref{fig_a3c_train}.
    \item Baseline algorithms where $\mathcal{F}$ could be fine-tuned to the source task were able to achieve higher asymptotic performance. This was particularly notable on DQN and A3C (see Figs.~\ref{fig_dqn_train} and ~\ref{fig_a3c_train}) in DeepMind Lab. However, in both those cases, DARLA was able to learn very reasonable policies on the source task which were on the order of 20\% lower than the fine-tuned models -- arguably a worthwhile sacrifice for a subsequent median 270\% improvement in target domain performance noted in the main text.
    \item Allowing DARLA to fine-tune its vision module (DARLA$_\text{FT}$) boosted its source task learning speed, and allowed the agent to asymptote at the same level as the baseline algorithms. As discussed in the main text, this comes at the cost of significantly reduced domain transfer performance on A3C. For DQN, however, finetuning appears to offer the best of both worlds.
    \item Perhaps most relevantly for this paper, even if solely examining source task performance, DARLA outperforms both DARLA$_\text{ENT}$ and DARLA$_\text{DAE}$ on both asymptotic performance and data efficiency -- suggesting that disentangled representations have wider applicability in RL beyond the zero-shot domain adaptation that is the focus of this paper.
\end{enumerate}

\end{document}